\documentclass{article}

\usepackage{todonotes}
\usepackage[export]{adjustbox}
\usepackage{amsfonts}

\usepackage{mathtools}
\usepackage{kotex}
\usepackage[ruled, linesnumbered]{algorithm2e}
\usepackage{multirow}
\DeclarePairedDelimiterX\set[1]\lbrace\rbrace{#1}
\newcommand{\squeezeup}{\vspace{-3mm}}

\usepackage{algorithmic}

\usepackage{hhline}
\usepackage{tabularx}
\usepackage{subfigure}
\usepackage{kotex-logo} 
\usepackage{setspace}
\usepackage{ textcomp }
\usepackage{upgreek}
\usepackage{xcolor}
\usepackage[toc,title,page]{appendix}
\usepackage{pifont}
\usepackage{color, colortbl}
\definecolor{brightgreen}{rgb}{0.4, 1.0, 0.0}
\definecolor{LightCyan}{rgb}{0.88,1,1}
\definecolor{Pink}{rgb}{0.94,0.57,0.61}
\definecolor{Gray}{gray}{0.9}
\newcommand*\colourcheck[1]{%
  \expandafter\newcommand\csname #1check\endcsname{\textcolor{#1}{\ding{52}}}%
}
\newcommand*\colourx[1]{%
  \expandafter\newcommand\csname #1x\endcsname{\textcolor{#1}{\ding{55}}}%
}
\colourcheck{black}
\colourcheck{green}
\colourcheck{red}
\colourcheck{brightgreen}
\colourx{black}
\colourx{green}
\colourx{red}
\makeatletter
\newcommand{\thickhline}{%
    \noalign {\ifnum 0=`}\fi \hrule height 1.2pt
    \futurelet \reserved@a \@xhline
}
\usepackage{textcomp}
\usepackage{gensymb}
\usepackage{svg}
\usepackage{dsfont}

\SetCommentSty{mycommfont}

\SetKwInput{KwInput}{Input}                % Set the Input
\SetKwInput{KwOutput}{Output}              % set the Output

\usepackage{graphicx}
\usepackage{amsmath}
\usepackage{amssymb}
\usepackage{booktabs}
\usepackage{wrapfig}

\usepackage[preprint]{corl_2022} % Uncomment for pre-prints (e.g., arxiv); This is like ``final'', but will remove the CORL footnote.

\usepackage{hyperref}
\hypersetup{
    colorlinks = true,
    linkbordercolor = {white},
    urlcolor = {magenta},
    citecolor = {green}
}
\title{Topological Semantic Graph Memory \\for Image-Goal Navigation} 

\author{
  Nuri Kim, Obin Kwon, Hwiyeon Yoo, Yunho Choi, Jeongho Park, and Songhwai Oh\\
  Department of Electrical and Computer Engineering, ASRI, Seoul National University \\
  \texttt{\{firstname.secondname\}@rllab.snu.ac.kr, songhwai@snu.ac.kr} \\
  \\
Webpage: {\small\texttt{\href{https://bareblackfoot.github.io/TopologicalSemanticGraphMemory}{https://bareblackfoot.github.io/TopologicalSemanticGraphMemory}}}
}

\begin{document}
\maketitle
%%%%%%%%% ABSTRACT
\squeezeup\squeezeup\squeezeup
\begin{abstract}
A novel framework is proposed to incrementally collect landmark-based graph memory and use the collected memory for image goal navigation. 
Given a target image to search, an embodied robot utilizes semantic memory to find the target in an unknown environment.
% 
% The semantic graph memory is collected from a panoramic observation of an RGB-D camera without knowing the robot's pose.
% 
In this paper, we present a topological semantic graph memory (TSGM), which consists of 
(1) a graph builder that takes the observed RGB-D image to construct a topological semantic graph, 
(2) a cross graph mixer module that takes the collected nodes to get contextual information, and 
(3) a memory decoder that takes the contextual memory as an input to find an action to the target.
On the task of an image goal navigation, TSGM significantly outperforms competitive baselines by +5.0-9.0\% on the success rate and +7.0-23.5\% on SPL, which means that the TSGM finds efficient paths. 
Additionally, we demonstrate our method on a mobile robot in real-world image goal scenarios.
% Using a real jackal robot, it was demonstrated that our method operates successfully s the real environment.
% 
% Code and Project page are available at \href{https://github.com/rllab-snu/TopologicalSemanticGraphMemory}{https://github.com/rllab-snu/TopologicalSemanticGraphMemory}.
\end{abstract}

\section{Introduction}
Navigation with rich visual observations has been a critical issue in a variety of embodied agent tasks, such as exploration, image goal navigation, and object goal navigation~\cite{mirowski2018learning, chen2019behavioral, gordon2019splitnet, yang2019embodied, Exp4nav, vaswani2017attention, kumar2018visual, SMT, SPTM, gupta2017cognitive, parisotto2017neural, henriques2018mapnet, avraham2019empnet, lv2020improving, objectGraphNavi, qiu2020target}. 
A crucial ingredient for successful visual navigation is to construct a $\textit{memory}$, which can represent the structure of the environment along with compact visual features for representing high-dimensional visual inputs.
A metric-map memory~\cite{ANS, Exp4nav} created with SLAM, and a graph memory~\cite{VGM, NTSLAM, SPTM, SMT, NeuralPlanner} with nodes and edges are the two standard memory construction approaches for navigation algorithms.
Even though navigation systems that use metric maps produce powerful results with exact localization and mapping, it is not practical because the navigation agent is susceptible to sensory noises.
The topological map, which represents geometric properties and spatial relations of places in the form of a graph, is proposed to construct a map without accurate mapping.
Previous visual navigation methods~\cite{SPTM, VGM} with topological map exploit image features as nodes and edges connecting the nodes in proximity. 
{\color{black}Since a node indicates a location, the robot's position can be estimated by the nodes in the topological map.}
% 
% Even if the agent's position is not precisely given, the agent might localize its position leveraging the nodes.
% 
Therefore, the graph can be used to navigate successfully even in a real noisy environment.

\begin{figure}[t!]{\centering\includegraphics[width=0.9\linewidth]{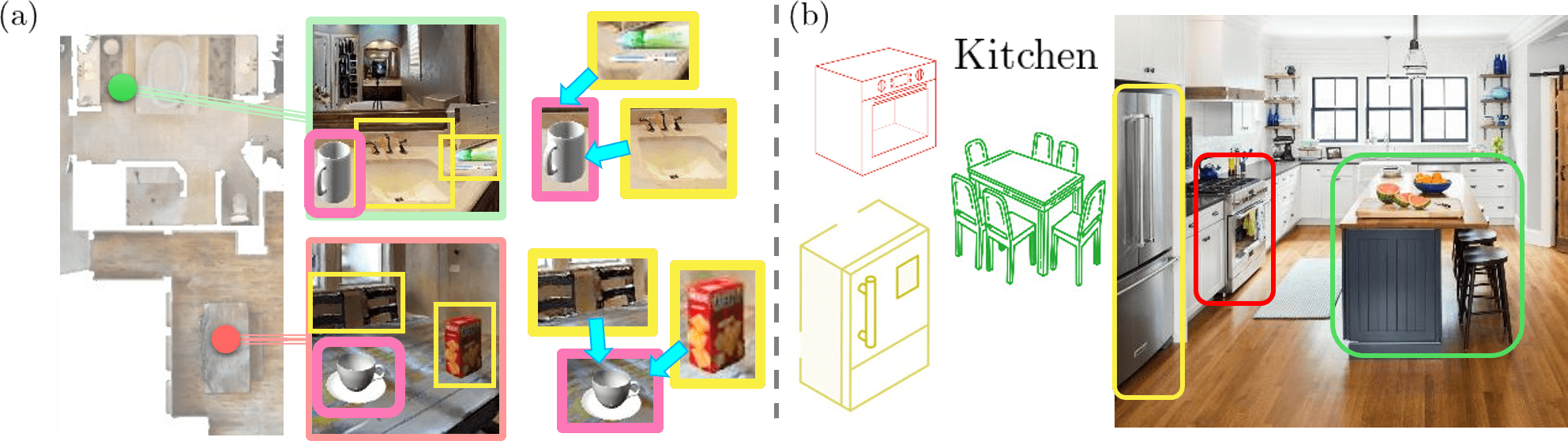}}\centering
\caption{{\textbf{\color{black}Importance of semantic contextual information.} (a) When two similar items, e.g., cups, are in different locations, they can be identified individually. (b) A kitchen can be recognized when refrigerator, oven, and dining table are detected.}
} \label{fig:overview}
\end{figure}

Unfortunately, using semantic information in topological graph memory presents several difficulties.
The first challenge is to incorporate landmarks into a topological graph.
As demonstrated by studies~\cite{landmarkBiology, chan2012objects, collett2004animal} that show animals navigate utilizing contextual cues from landmarks, the problem of incorporating landmarks, such as object compositions, is a critical issue in visual navigation.
A recent research~\cite{objectGraphNavi} addresses the problem by making an object graph, which connects objects in a field of view of a directional camera to leverage object features.
Although it guides an agent to effective action to discover the target utilizing object relationships, it often misses 3D object context information since it only connects objects observed from the same viewpoint.
% 
% Moreover, it do not construct an explicit memory to store the past trajectory.
% 
Imagine a goal is given to find a cooking pot, and you know that the cooking pot is usually kept adjacent to an oven. 
If the oven is nearby but not visible, the agent may miss this crucial information and pick an inefficient path.
The more difficult problem is inferring contextual information from objects using geometrically arranged landmarks.
% 
% In other words, connecting the object graph is critical problem.
% % 
% The recent method~\cite{objectGraphNavi} only connects the objects in the image.
% % 
% It does not represent the genuine spatial relationship of objects.
% 
There are various advantages to obtaining contextual features from topological graphs.
By defining an object through neighboring objects, the contextual representation helps to eliminate the ambiguity of similar but different objects.
% 
% As shown in Figure~\ref{fig:overview}(a), a cup in the kitchen and a cup in the bathroom can be distinguished by updating the cup in the kitchen with nearby objects such as a dining table and an oven, while the cup in the bathroom is updated with a toothbrush.
As shown in Figure~\ref{fig:overview}(a), {\color{black}by describing cups in terms of their neighboring objects, it is possible to distinguish two similar cups.
For example, a cup in the kitchen can be perceived as one next to a chair and snack box, while a cup in the bathroom can be shown as one that is near to a toothbrush and washstand.}
% 
% The green path is for the cup in a bathroom, and the red path is for the cup on a kitchen table. 
% 
% When the objective is to find a cup from a kitchen, an agent can chooses to follow the red path.
% 
Moreover, a place can be better described through objects.
A kitchen, for example, can be defined by the presence of a refrigerator, oven, and dining table (see Figure~\ref{fig:overview}(b)).
% 
% In the previous methods~\cite{objectGraphNavi}, object context information was concatenated to the observed image of the object, which is helpful in that the object context defines a certain space, but does not make objects have individuality.

To this end, this paper addresses the challenges described above using topological semantic graph memory (TSGM), which has two types of nodes: image nodes and object nodes.
Image nodes represent different positions that include image representations, while each object node represents a unique object with its visual representation regardless of viewpoint.
If a previously undiscovered image node is identified, it is connected to the previously visited image node, producing a spatially meaningful graph.
For each image input, objects are detected using MaskRCNN~\cite{he2017mask} and then connected to image nodes where the object can be seen.
% 
% Then, the objects connected to the same image node or the image nodes neighboring the current image node are geometrically proximate.
% 
A cross graph mixer, a learnable message-passing network that exchanges object and image information, is then used to make the memory contextual.
% 
% Self-update(object update / image update) 와 cross update 로 되어있다고 설명하기
% 
Using this contextual memory, the agent determines the best strategy for finding the goal.
We applied our proposed method to image goal navigation in the Gibson \cite{GibsonEnv} to validate our method. 
% 
% As a result, our method shows state-of-the-art performance, resulting in 82.5\% success rate and 74.1\% success rate over path length (SPL) on curved episodes.
As a result, the proposed method significantly outperforms competitive baselines by +5.0-9.0\% on the success rate and +7.0-23.5\% on SPL, which means that the TSGM finds efficient paths.
% 
% It means that the TSGM successfully finds a more efficient path to reach the target.
% 
On a mobile jackal robot, we demonstrate our method in a real-world environment. 
The experiment is carried out using episodes where the goal location is not visible and is 5 to 10m away. 
Despite that the episodes are challenging for image goal navigation, TSGM demonstrates successful navigation.

\section{Related Work}
There has been a lot of research using memory for visual navigation.
There are three types of memory formation methods: implicit memory, metric memory, and topological memory.

\paragraph{Implicit memory.} 
Using a recurrent neural network (RNN) as a policy network is a simple method to make an implicit memory~\cite{TargetDriven}. 
TargetDriven~\cite{TargetDriven} has a vanilla RL policy with a CNN backbone followed by an LSTM which is implicit memory.
% 
% The hidden state in the LSTM network is implicit memory.
%  
Since RNN has difficulty backpropagating for a long sequence, RNN is replaced with an explicit memory structure~\cite{ANS, Exp4nav, SMT, chen2019behavioral, NeuralPlanner, SPTM}. 

\paragraph{Metric memory.} 
{\color{black}Active Neural SLAM}~\cite{ANS} has a hierarchical structure to explore an environment: global and local policy. 
The global policy constructs a top-down 2D map and estimates a global goal. 
It consists of a neural network for flexible output on the input modalities. 
Given the global goal from the global policy module, a local policy module plans a path to the goal using the simple local navigation algorithm.%, such as Fast Marching Methods~\cite{FMM}. 
Exp4nav~\cite{Exp4nav} tackles the exploration problem for navigation. 
It builds a global metric map, combining egocentric metric maps. 
Then, to estimate the action, it embeds images with CNN, and a recurrent policy takes the embedding of current observation and the target to output an action. 
% 
% It firstly trains an agent with imitation learning and finetunes the network with reinforcement learning with a coverage reward. 
% 
% MapNet~\cite{henriques2018mapnet} and EMPNet~\cite{avraham2019empnet} can be used for navigation, even though they are not made for navigation. 
% 
% MapNet~\cite{henriques2018mapnet} and EMPNet~\cite{avraham2019empnet} use RGB images and depth images to embed them into memory in the form of spatial maps for localization.

\paragraph{Topological memory.}  
In \cite{chen2019behavioral}, LSTM is a policy network that finds the currently located node and derives an action when a topological map is provided. 
% 
% \cite{chen2019behavioral} uses imitation learning to train the network.
% 
The goal of NeuralPlanner~\cite{NeuralPlanner} is primarily to explore the environment and maximize coverage. 
To this end, a topological map is generated from an exploratory policy during the rollout. 
Then, a pretrained neural planner calculates the path to the node most similar to the target image. 
Semi-parametric topological memory (SPTM)~\cite{SPTM} forms a topological memory through exploration before training an agent.  
% 
% SPTM contains a non-parametric graph memory where each node means a location of a map and a parametric retrieval network. 
% 
% SPTM first forms a topological memory through exploration before training an agent. 
% 
Then, an agent navigates to the destination based on the topological memory using the Dijkstra algorithm that plans a path to reach the waypoint.
{\color{black}Scene memory transformer (SMT)}~\cite{SMT} stacks all the visual features of the past observations as a navigation memory. 
It uses transformer network~\cite{vaswani2017attention} to process this memory in the context of the current and target observations.
In other words, the authors completely replaced the RNN with an attention mechanism, which shows good performance in long-term tasks.
{\color{black}Visual graph memory (VGM)}~\cite{VGM} constructs a topological visual memory to navigate an environment while not utilizing the landmark information of the scene. 
{\color{black}No RL no simulator (NRNS)}~\cite{NRNS} uses models trained using image input without interaction with the simulator. 
It creates a node for an unexplored area and finds the goal by estimating the geodesic distance to the goal.

Our method builds upon prior topological memory methods to handle the problem of creating and exploiting the object-based semantic graph to find the most efficient path.
Once placed in a new environment, our method explores while incrementally constructing a topological graph, inspired by prior work~\cite{VGM, SMT, NRNS}.
In contrast to these previous methods, our method employs spatially meaningful landmarks without using exact positions.

\begin{figure*}[t!]{\centering\includegraphics[width=0.9\linewidth]{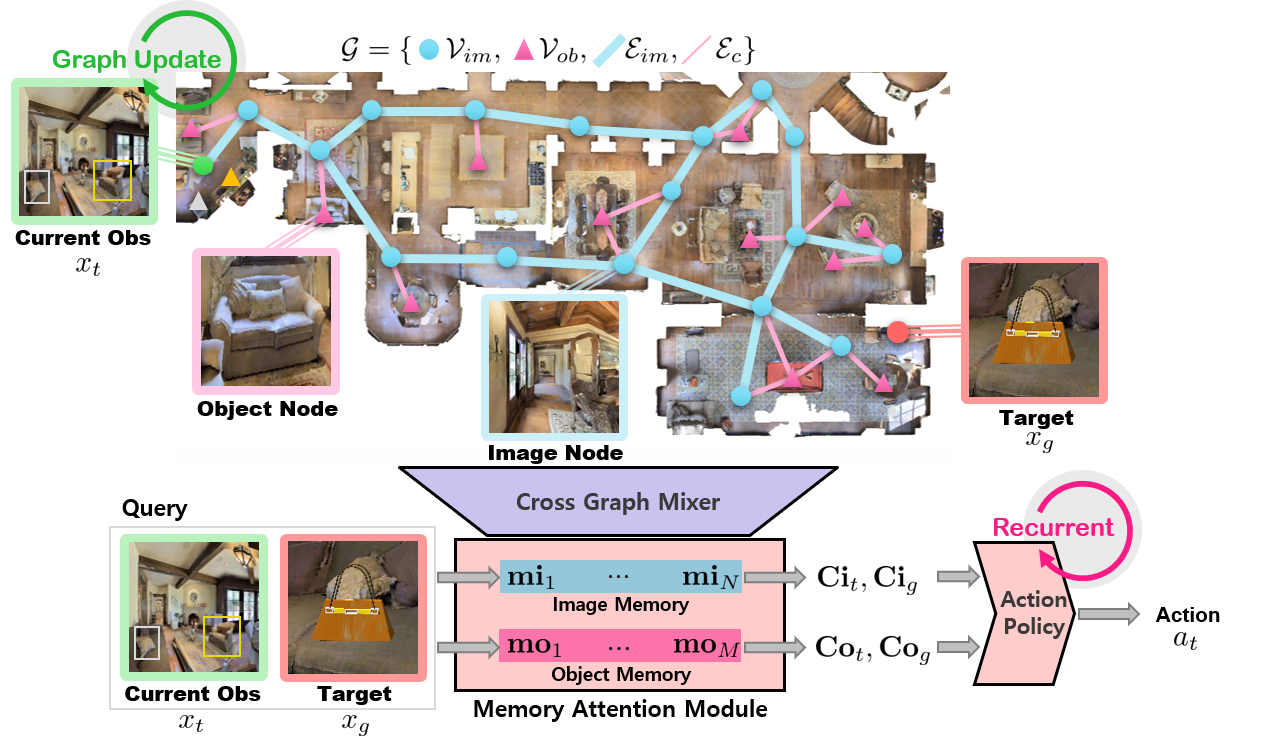}}\centering
\caption{\textbf{Overview of the proposed method.}
TSGM has a spatially meaningful structure that is generated and updated online.
An image node $\mathcal{V}_{im}$ is represented by a blue circle, and an object node $\mathcal{V}_{ob}$ is represented by a pink triangle. 
The current observation $x_t$ is shown in green, while the goal $x_g$ is highlighted in red.
A cross graph mixer module is used to update the constructed graph memory to encode the context of features. 
Then, the attention module chooses the updated memory by querying the current observation and target node.
The action policy network determines the action $a_t$ based on the selected memory.
{\color{black}Note that the node position is only used for visualization purposes.}
} \label{fig:network} 
\vspace{-10pt}
\end{figure*} 

% \section{Proposed Method} ~\label{sec:tsgm}
\section{TSGM: Topological Semantic Graph Memory}
\subsection{Problem Statement}
We consider the problem of goal-directed exploration for efficient navigation.
Given a current observation and a target image to search, an embodied robot aims to find the target (see Figure~\ref{fig:network}). 
% 
% Let us consider a mobile agent that is equipped with an RGB-D camera and that can execute basic movement macro-actions.
% 
The current observation contains RGB-D sensory input $x_t$ and detected objects $\{z_1, ..., z_K\} \in \textbf{z}_t$.
% 
% {\color{black}
% The current observation contains RGB-D sensory input $I_t$ and detected objects $\textbf{z}_t$.
% }
% 

% We model the task of navigation as a Markov decision process.
% \subsection{Problem Statement}
\subsection{Graph {\color{black}M}emory {\color{black}C}onstruction}
% \subsection{TSGM: Topological Semantic Graph Memory}~\label{sec:tsgm}
Let us consider an agent equipped with an RGB-D camera that was dropped into a novel environment that it has never been in before. 
% 
% This agent has been dropped into a novel environment that it has never been in before. 
% 
We want to build and exploit a {\color{black}graph memory} that enables this agent to find the goal in a new environment efficiently.
% We want to build and exploit a topological semantic graph memory that enables this agent to find the goal in a new environment efficiently.
% 
To this end, a semantically structured graph memory, named topological semantic graph memory (TSGM), is built online while an agent traverses the new environment.
For example, as shown in Figure~\ref{fig:network}, when a new node (green circle) is found, the new node is connected to the previous image node (the blue circle connected to the green circle), and the object nodes found at the new node (the green triangles). 
Since the graph is built with landmarks, recalling the landmarks the agent has seen before will help the agent to discover efficient paths.
% 
% For example, if you put the oven in the object memory to find a place with a cooking pot, it is easier to infer that a cooking pot is close.
% 
% \paragraph{Graph memory construction.}
The TSGM contains two types of nodes and edges, i.e., $\mathcal{G} = \{\mathcal{V}_{im}, \mathcal{V}_{ob}, \mathcal{E}_{im}, \mathcal{E}_{c}\}$ with image nodes $\mathcal{V}_{im} = \{x_i\}^N_{i=1}$ and its edges $\mathcal{E}_{im}$ with the affinity matrix $A_{im} \in \mathbb{R}^{N \times N}$, object nodes $\mathcal{V}_{ob} = \{z_i\}^M_{i=1}$ where $z_i \in \mathbb{R}^{1 \times D}$, and edges connecting image and object nodes $\mathcal{E}_c$ with the affinity matrix $A_{c} \in \mathbb{R}^{N \times M}$. 
$N$ is the number of image nodes, $M$ is the number of object nodes, and $D$ is the dimension of the object representation.
The graph expands continuously as the agent explores the environment.
{\color{black}The algorithm for constructing TSGM is provided in Appendix~\ref{sec:sup2}.}

\paragraph{Object graph construction.}
Objects are retrieved using MaskRCNN~\cite{he2017mask} and added to the graph memory as nodes.
When the object is already in memory, the object node is updated with a node with a higher object detection score.
If it is not determined that the object is the same, a new object node ($z_k$) is added to the memory, connecting with the corresponding current image node ($x_t$) using the affinity matrix ($A_{c}$). %  $A_{c}[x_t,z_k]$ = 1, which is the image-object affinity matrix of TSGM.
% 
% To reduce computational complexity, we only check the similarity of the connected object nodes.
% 
To connect objects in proximity in the graph memory, we calculate the affinity matrix for object nodes to be connected to objects in the neighbors of the current image node using the image affinity and image-object affinity.
Here, since we want a single object node for an object, the trained object encoder using Supervised contrastive learning~\cite{khosla2020supervised} learns whether the objects are the same object even when the images are shot from various points of view.
% 
% For this, Supervised contrastive learning~\cite{khosla2020supervised} is used to learn the same object to be closer in metric space.

\paragraph{Image graph construction.}
To construct an image graph, a pretrained similarity encoder~\cite{li2021prototypical} learns image similarity in an unsupervised manner and is then used to determine whether or not the node is already in memory.
% 
% The pretrained similarity encoder is used to determine whether or not the node is already in memory.
% 
If an observed node is new, i.e., the node feature is not similar to the existing memory, it is added to the graph memory; otherwise, it is utilized to update the corresponding memory node.
While connecting image nodes, a new technique based on objects is utilized, assuming that the likelihood of being at the same place is significantly low if there are few co-visible objects between two image nodes.
Therefore, the similarity between the object nodes collected at the current location and the object graph nodes is calculated.
We consider two image nodes to represent different locations when there are no similar items between them.
If the similarity is low, the observed image node is added to memory; otherwise, the image node is updated to the most recent image representation.
When it turns out that the image node is new, it is connected to the previous image node ($x_l$), making the image affinity matrix  ($A_{im}$) between $x_l$ and $x_t$ connected.
% 
% After the image graph and the object graph are updated respectively, the image node is updated as connected object nodes, and the object nodes are updated using the connected image nodes. For the best view, the goal feature appended to each node is omitted, and only $v$th node, which is a node to be updated, in the $l$th layer is visualized.
% 

\subsection{Cross Graph Mixer}   
% \begin{wrapfigure}{o}{0.7\textwidth}
\begin{figure}[t!]
{\centering\includegraphics[width=0.7\linewidth]{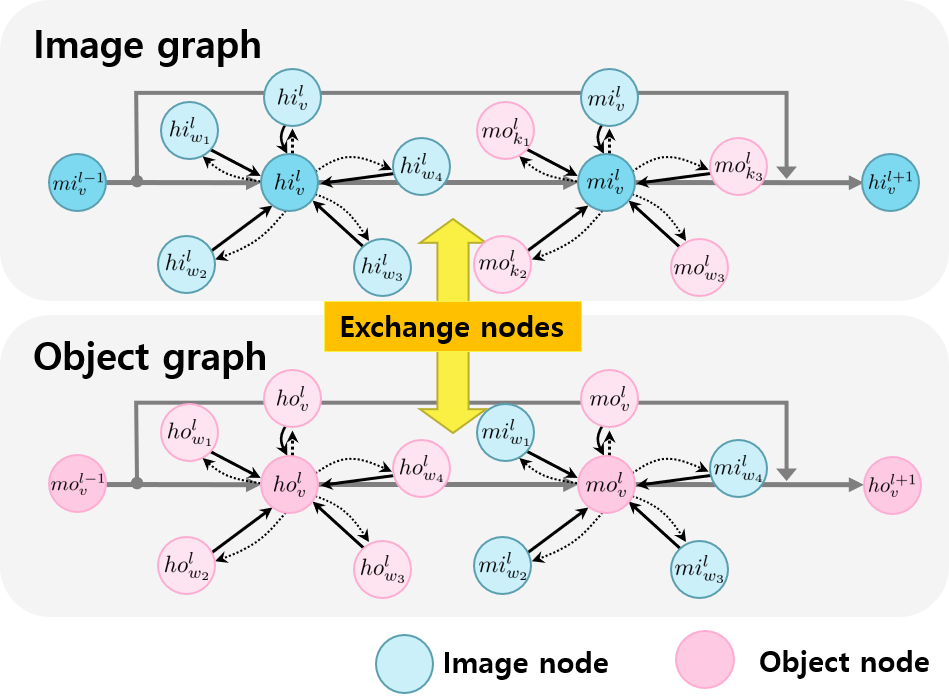}}\centering
\caption{\textbf{Cross graph mixer.} 
% The image graph update (upper) and object graph update (lower) work in parallel.
The upper row shows the image graph update while the lower row shows the object graph update.
} \label{fig:crossupdate}
\end{figure} 
% \end{wrapfigure}
% 
The cross graph mixer is developed from MPNNs~\cite{MPNN} to encode scene contexts by combining information from the image and object graphs.
The upper row of Figure~\ref{fig:crossupdate} depicts the image graph, while the lower row represents the object graph.
The message passing phase runs for $L$ steps and is defined of message functions $M^l$ and vertex update functions $U^l$, where $l \in \{1, ..., L\}$. The message function $M^l$ is composed of two functions, the self-update function ($S^l$) and the cross-update function ($C^l$). %$M^l = C^l \circ S^l$, 
To begin, image and object nodes self-update to obtain contextual representations of nearby locations or objects, respectively:
\begin{align}
\begin{aligned}{
    \hat{mi}_v^{l} = \sum_{w \in \mathcal{N}_i(v)} S_{i}^l({hi}_w^l, A_{im}, g), \;
    \hat{mo}_v^{l} = \sum_{k \in \mathcal{N}_o(v)} S_{o}^l({ho}_k^l, A_{ob}, g)},
\end{aligned}
\end{align}
where $\mathcal{N}_i(v)$ and $\mathcal{N}_o(v)$ denote the image/object neighbors of the $v$th node. 
The node features $h$ are concatenated with the goal feature $g$, and it is embedded into a feature using two-layered neural networks. 
Then, the self-update message function $S$ makes $\hat{mi}_v^{l}$ and $\hat{mo}_v^{l}$ by aggregating connected nodes with the message passing method, which is illustrated in the left column of Figure~\ref{fig:crossupdate}.
After self-update, each graph exchanges information with the others to make a complete the messages,  %
\begin{align}
\begin{aligned}{
    {mi}_v^{l} = \sum_{k \in \mathcal{N}_o(v)} C_i^l(\hat{mi}_v^l, \hat{mo}_k^l, A_{c}), \;
    {mo}_v^{l} = \sum_{w \in \mathcal{N}_i(v)} C_o^l(\hat{mo}_v^l, \hat{mi}_w^l, A_{c})}, 
\end{aligned}
\end{align}
where $C_i$ and $C_o$ are cross updating functions that produce ${mi}_v^{l}$ and ${mo}_v^{l}$, which are cross-updated messages.
The cross-update module's central concept is that images are updated through connected objects, and objects are updated using connected images.
To cross-update the graphs, the self-updated image/object features aggregate the connected object/image node features.
% 
% It indicates that the objects receive connected image node features, whereas the image nodes receive connected object node features.
% 
Then, the update function $U$ transfers messages from object nodes to image nodes and vice versa,
\begin{align}
\begin{aligned}{
    {hi}_v^{l+1} = U_i^l({hi}_v^l, {mi}_v^{l}),\;
    {ho}_v^{l+1} = U_o^l({ho}_v^l, {mo}_v^{l})}.
\end{aligned}
\end{align}
where $U$ is a two-layered neural network that maps information of an object/image to the connected objects/images.
The image-object update iterates for $L$ steps to create semantic contextual node features.
After $L$ steps of the image-object update iterations, the cross-mixed memory for image graph $\textbf{mi}$ and object graph $\textbf{mo}$ are generated for each node.
The details of the module structure are described in the Appendix~\ref{sec:sup3}.

% \section{Memory Attentional Policy}
\subsection{Memory Attention Module}~\label{sec:pap3.4}
TSGM is composed of visited image nodes and object nodes that have been observed.
For this, an attention network is used to discover the node closest to the goal among the memory.
Since this module utilizes attention, interpolation between visited nodes enables goal features to be extracted from unexplored nodes.
Additionally, to find the relative path, the current node is better to be searched.
The goal feature $x_g$ is given as a query $q$ when extracting a memory-conditioned goal feature,  $\textbf{Ci}_g$, $\textbf{Co}_g$, and the current observation feature $x_t$ is given as an input when selecting the memory-conditioned current feature, $\textbf{Ci}_t$, $\textbf{Co}_t$.
% 
% This process is carried out for each image graph and object graph.
% are combined and used to select an action.
% 
Using the decoder module of the transformer network~\cite{vaswani2017attention}, current contextual feature $\textbf{Ci}_t$, $\textbf{Co}_t$ and goal context feature $\textbf{Ci}_g$, $\textbf{Co}_g$ are collected,
\begin{align}
\begin{aligned}{
\textbf{Ci} =  \sigma\Big{(}\frac{(\textbf{W}_qq)(\textbf{W}_k{\{\textbf{mi}_1, ..., \textbf{mi}_n\}^T)}}{\sqrt{d}}\Big{)}(\textbf{W}_vq),
\textbf{Co} =  \sigma\Big{(}\frac{(\textbf{W}_qq)(\textbf{W}_k{\{\textbf{mo}_1, ..., \textbf{mo}_n\}^T)}}{\sqrt{d}}\Big{)}(\textbf{W}_vq)},
\end{aligned}
\end{align}
where $\textbf{W}_v$, $\textbf{W}_q$, and $\textbf{W}_k$ are matrix parameters and the set ${\{\textbf{mi}_1, ..., \textbf{mi}_N\}}$ is the image memory, the set ${\{\textbf{mo}_1, ..., \textbf{mo}_M\}}$ is the object memory, and $\sigma$ is a softmax function. 

\subsection{Action Policy Network}
Given the contextual feature for goal and current nodes, the action policy network finds an action to reach the goal.
For $t$th time step, we encode the sequential information, $\textbf{h}_a^t = G(\textbf{h}_c^t, \textbf{h}_g^t, \textbf{h}_a^{t-1}, a_{t-1}),$ where $G$ is a recurrent neural network and $\textbf{h}_a^t$ is a $t$th hidden state for action, and $a_{t-1}$ is the previous action of an agent. 
Then, the action $a_t$ is sampled from the Categorical distribution, i.e. $a_t \sim$ Categorical($\mathbf{W}\textbf{h}_a^t + \mathbf{b})$, where $\mathbf{W}$ and $\mathbf{b}$ are matrix parameters.

\subsection{Optimization}
\paragraph{Learning action.}
Proximal policy optimization (PPO)~\cite{PPO} is used to learn a policy for picture goal navigation.
The policy is trained to maximize the expected return, which is defined as the total reward $\mathbb{E}_\tau[R(s_t,a_t)]$ over time trajectories {\color{black} $\tau = (s_t,a_t)^H_{t=1}$} of the policy's time horizon $H${\color{black}.}
When an agent takes action $a_t$ at state $s_t$, $\mathcal{L}_{ppo} = -\mathbb{E}_t \Big{[} {\nabla_\theta\text{log} \pi_\theta(a_t|s_t) \hat{A}_t}\Big{]},$ where and $\hat{A}_t$ is an estimator of the advantage function at time step $t$, and $pi theta$ is a stochastic policy.
We formulated it as a negative log likelihood to the oracle action, 
$\mathcal{L}_{act} = -\mathbb{E}_{\tau \sim \mathbb{D}} \Big{[} \sum_{t=1}^{T_{\tau}}{{a}^*_t \text{log}(a_t|x_t)}\Big{]},$ where $\tau$ = $(x_t, a_t^*)$ for $t \in \{1,2,...,T_{\tau}\}$, and ${a}_t^*$ is the oracle action at time step $t$.
Since PPO can better estimate return after creating successful episodes, the policy is pretrained with imitation learning and then finetuned with reinforcement learning.

\paragraph{Auxiliary metrics.}
Knowing when to stop is a significant issue since an episode is considered successful if an agent calls the stop action upon seeing the goal.
Fortunately, it is well known that auxiliary metrics, such as a progress monitor and goal sensor, can assist an agent in determining when to stop~\cite{RedRabbit}.
% 
% The improvement on better stopping is accomplished by using two auxiliary metrics: progress monitor and goal sensor.
% 
The progress monitor assesses progress based on the current observation and a target to determine when to press the stop button, which is obtained by calculating the geodesic distance between an agent and the goal.
% and then utilizing the overall geodesic distance of an episode to normalize it.
% 
The goal sensor indicates how close the current position is to the goal.
% 
% When using the goal sensor for the object sensor, the target object's presence in the current observation triggers the sensor.
% 
If the target is within the success criteria for image goal navigation, the goal sensor produces one.
The progress monitor takes the current observation and a target as input and estimates the progress, $\hat{p}_{t}$.
% i.e. $\hat{p}_t$ = $f_p(x_t, g)$, where $f_p$ is a neural network. 
% 
The target sensor takes the same input of the progress monitor and estimates that the current observation containing a target, $\hat{s}_{t}$. % $\hat{s}_{t}$ = $f_s(x_t, g)$, where $f_s$ is a neural network. 
Then, two estimations, $\hat{p}_t$ and $\hat{s}_{t}$, are optimized with $\text{L}_2$ loss,
$\mathcal{L}_{aux} = \mathbb{E}_{\tau \sim \mathbb{D}} \Big{[} \sum_{t=1}^{T_{\tau}} ||s^*_t - \hat{s}_t||_2 + ||p^*_t - \hat{p}_t||_2  \Big{]},$ where $s^*_t$ and $p_t^*$ are the ground truths. 
Since the auxiliary loss is simply added to the action loss, the final loss function for training an agent with imitation learning is $\mathcal{L}_{bc} = \mathcal{L}_{act} + \lambda \mathcal{L}_{aux}$, where $\lambda$ is a balancing parameter. 
For reinforcement learning, the combined loss $\mathcal{L}_{rl} = \mathcal{L}_{ppo} + \lambda \mathcal{L}_{aux}$, is used.

%%%%%%%%%%%%%%%%%%%%%%%%%%%%%%%%%%%%%%%%%%%%%%%%%%%%%%%%%%%%%%%%%%%%%%%%%%%%%%%%
\section{Experimental Evaluation}
% 왜 undirected camera 쓰는지 설명 (실제로 neuron이 rotation에 영향을 안받는다는 사실...)
% By comparing our algorithm with other visual navigation methods, we get compelling results for the task of path following both in simulations and real robot experiments. The robot experiments are done with a Jackal robot and the navigation algorithms run on a notebook running Ubuntu 16.04 with a 2.6GHz Intel Core i7-9750H CPU and an NVIDIA GeForce RTX 2070 and the experiments on the simulator use a server with Ubuntu 18.04, a 2.4GHz Intel Xeon E5-2680 CPU and an NVIDIA Titan RTX.

\subsection{Baselines} 
We compare our method with baselines with various types of memory.% for image goal navigation. 
\textbf{RGBD + RL}~\cite{TargetDriven} has a vanilla RL policy with a CNN backbone followed by an LSTM adapted from \cite{TargetDriven}.
\textbf{\color{black}Active Neural SLAM}~\cite{ANS} has a metric memory for exploration. To adapt the algorithm to the image goal navigation, we set the output of the global policy to be the relative position to the target when the target is detected using a pretrained target pose estimator. 
\textbf{Exp4nav~\cite{Exp4nav}} tackles the exploration problem.% for navigation while building a metric map using exact pose information. 
To finetune the image goal navigation task model, we change the coverage reward to the image goal reward.  
\textbf{Neural Planner}~\cite{NeuralPlanner} is a model adapted the method~\cite{NeuralPlanner} for image goal navigation from exploration task. % discovers a waypoint rather than a destination node. Before training an agent, it 
\textbf{SPTM}~\cite{SPTM} creates a topological graph, and then the Dijkstra algorithm then creates a path to a waypoint.
\textbf{SMT}~\cite{SMT} stacks all the visual features of the past observations and the pose information as a navigation memory.
\textbf{VGM}~\cite{VGM} builds a topological memory to navigate an environment without using landmark information.
\textbf{NRNS}~\cite{NRNS} trained agent without interaction with the simulator. In order to compare the method fairly, we adapted the method to use a panoramic camera.

\subsection{Experiment Settings} 
% 실제로 돌아다니다가 물체의 구성이 변화된 것에 정말 robust한지 측정하는 실험이 있으면 좋을듯
We evaluate TSGM in the Gibson environment with a habitat simulator, which is photo-realistic.
For image observation, we used a panoramic image, the same setting with~\cite{VGM}, inspired by the human neuron not recognizing the heading for localizing~\cite{placeBiology}.
% 
% For detecting objects, Mask RCNN~\cite{he2017mask} trained on the COCO dataset~\cite{lin2014microsoft} for the panoptic segmentation with ResNet101 backbone is used to detect objects.
% 
The ground truth objects from the Gibson dataset and detected objects from the detector are used for training. For testing, a detector~\cite{he2017mask} pretrained with a COCO dataset is used. 
We use a discrete action space defined as $\mathcal{A} =\{\texttt{go forward}, \texttt{turn left}, \texttt{turn right}, \texttt{stop}\}$, which is a common choice for navigation problems. 
The step size of the forward action is 0.25$m$, and the rotation angle is set to $10\degree$ in all experiments.
The reward function $R$ is defined using the progress of an agent, using a geodesic distance between an agent and the target, and +10 is given when the agent reaches the goal.
%  $R = |d(t) - d(t-1)|$, where $d$ is
% 
{\color{black}Further implementation details of the proposed method are provided in Appendix~\ref{sec:sup1}.}

% \paragraph{Similarity thresholds.}  
% For constructing TSGM, we use a pretrained network to determine whether to add or update the graph with a new observation. 
% % 
% To this end, we set the similarity threshold for image $\tau_i$ to 0.75 as in VGM~\cite{VGM}. For object threshold, $\tau_o$, the distribution of the successful clustering on the test scenes are found after training object features with Supervised Contrastive Learning~{\cite{khosla2020supervised}}. 
% % 
% For training objectness, we used object id as a target. Then, we found a boundary with the largest margin for differentiating the two objects with different id, $\tau_o$ = 0.8. 
% % 
% The object encoder showed promising results in predicting object identifications on the clustering measures with 0.839 NMI and 0.878 purity.
% ==> supplementary

\textbf{Evaluation metrics.}~\label{eval_metric}
We use \textbf{Success} and \textbf{SPL} (success rate over path length) to evaluate the navigation tasks. 
\textbf{Success} is calculated by dividing the total success number by the number of test episodes. 
We set the success criterion as 1$m$ to the goal, a common criterion for image goal navigation.
\textbf{SPL} multiplies the success rate and the ratio of the shortest path length and traveled path. 
% When there are $N$ episodes, $\text{SPL} = \frac{1}{N} \sum_{i=1}^N S_i \frac{l_i}{\max(p_i, l_i)}$, where $l_i$ is the length of shortest path between goal and target, $p_i$ is the length of path taken by agent, and $S_i$ is the binary indicator of success in episode $i$.
% Since the ratio is smaller than 1, the SPL is always smaller or the same compared to the success rate.

\textbf{Episode settings.}
We divide test episodes into three difficulty levels for image goal navigation: easy, medium, and hard. 
The difficulty is determined by the geodesic length of an episode. 
Following \cite{ANS, VGM}, we set the length as follows: easy (1.5 $\textendash$ 3m), medium (3 $\textendash$ 5m), and hard (5 $\textendash$ 10m).
% For image goal navigation, We divides test episodes into three difficulty settings: {easy}, {medium}, and {hard}. The geodesic length of an episode is the criterion for determining the difficulty. Following~\cite{ANS, VGM}, we choose the episode length as follows: {easy} (1.5 $\textendash$ 3m), {medium} (3 $\textendash$ 5m), and (5 $\textendash$ 10m) for hard.
% Furthermore, we sample the start-goal pairs for episodes corresponding to the area size of the test scene since there are a lot of similar episodes duplicated if we generate episodes in a small scene.

\begin{table*}[!t]
\centering
\caption{{Comparison of TSGM with memory-based baselines on image goal navigation on Gibson.}}
{\resizebox{\linewidth}{!}{
\begin{tabular}{llcc|cc|cc|cc|cc} \toprule 
 \multirow{2}{*}{\textbf{Method}} &
 \multirow{2}{*}{\textbf{Memory}} &
 \multirow{2}{*}{\textbf{No Pose}} &
 \multirow{2}{*}{\textbf{Object}} & \multicolumn{2}{c}{\textbf{Easy}} & \multicolumn{2}{c}{\textbf{Medium}} & \multicolumn{2}{c|}{\textbf{Hard}} & \multicolumn{2}{c}{\textbf{Overall}} \\  \cmidrule{5-12}
& & & & \multicolumn{1}{c}{Success} & \multicolumn{1}{l}{SPL} & \multicolumn{1}{c}{Success} & \multicolumn{1}{c}{SPL} & \multicolumn{1}{c}{Success} & \multicolumn{1}{c|}{SPL} & \multicolumn{1}{c}{Success} & \multicolumn{1}{c}{SPL} \\ \midrule
RGBD + RL~\cite{TargetDriven} & implicit & \redx & \redx & 72.5 & 69.5 & 53.1 & 48.6 & 22.3 & 17.7 & 49.3 & 45.3 \\
{\color{black}Active Neural SLAM}~\cite{ANS} & metric & \redx & \redx & 74.2 & 20.5 & 68.4 & 22.9 & 29.9 & 11.0 & 57.5 & 18.1 \\
Exp4nav~\cite{Exp4nav} &  metric & \redx &  \redx & 70.2 & 61.8 & 60.6 & 52.4 & 46.9 & 38.5 & 59.2 & 50.9 \\
SMT~\cite{SMT}  & graph & \redx &  \redx & 81.9 & 77.4 & 65.6 & 52.2 & 55.6 & 39.7 & 67.7 & 56.4 \\
Neural Planner~\cite{NeuralPlanner} &  graph & \redx &  \redx & 71.7 & 41.3 & 64.7 & 38.5 & 42.0 & 27.0 & 59.5 & 35.6 \\
SPTM~\cite{SPTM} & graph & \brightgreencheck &  \redx & 66.5 & 40.6 & 64.2 & 38.5 & 42.1 & 25.4 & 57.6 & 34.8 \\
VGM~\cite{VGM} &  graph & \brightgreencheck & \redx & 86.1 & 79.6 & 81.2 & \textbf{68.2} & 60.9 & 45.6 & 76.1 & 64.5 \\ \midrule
\rowcolor{LightCyan}
TSGM (Ours) &  graph & \brightgreencheck & \brightgreencheck & \textbf{91.1} & \textbf{83.5} & \textbf{82.0} & 68.1 & \textbf{70.3} & \textbf{50.0} & \textbf{81.1} & \textbf{67.2} \\   \bottomrule
\end{tabular}}}
\label{tab:imagegoal_vgm}  
\end{table*}

\begin{table*}[!t]
\centering
\caption{{Comparison of TSGM with image goal navigation baselines on straight/curved episodes on Gibson.}}
{\resizebox{0.88\linewidth}{!}{
\begin{tabular}{c|l|rr|rr|rr|rr} \toprule 
\multirow{2}{*}{\textbf{Path Type}} & \multirow{2}{*}{\textbf{Method}}        &  \multicolumn{2}{c}{\textbf{Easy}}                                   & \multicolumn{2}{c}{\textbf{Medium}}                                 & \multicolumn{2}{c|}{\textbf{Hard}}                                   & \multicolumn{2}{c}{\textbf{Overall}}                                \\ \cmidrule{3-10} 
                             & & \multicolumn{1}{c}{Success} & \multicolumn{1}{l}{SPL} & \multicolumn{1}{c}{Success} & \multicolumn{1}{c}{SPL} & \multicolumn{1}{c}{Success} & \multicolumn{1}{c|}{SPL} & \multicolumn{1}{c}{Success} & \multicolumn{1}{c}{SPL} \\ \midrule
% \multirow{4}{*}{Straight}  & RL (10M step)                & 10.5          & 6.7    & 18.1           & 15.1    & 11.7          & 10.8   & 13.4            & 10.9    \\
                        %   & RL (extra data + 100M steps) & 43.2          & 38.5   & 36.4           & 34.8    & 7.4           & 7.2    & 29.0            & 26.8    \\
                        %   & BC w/ ResNet + Metric Map    & 24.8          & 23.9   & 11.5           & 11.2    & 1.3           & 1.2    & 12.5            & 12.1    \\
                        %   & BC w/ ResNet + GRU           & 34.9          & 33.4   & 17.6           & 17.0    & 6.1           & 5.9    & 19.5            & 18.8    \\
                        %   & NRNS~\cite{NRNS}                         & 68.0          & 61.6   & 49.1           & 44.5    & 23.8          & 18.2   & 46.9            & 41.4    \\
\multirow{3}{*}{Straight}                           & NRNS~\cite{NRNS}                        &  67.1 & 57.8 & 52.4 & 41.2 & 32.6 & 22.4 & 50.7 & 40.5     \\
                          & VGM~\cite{VGM}                          & 81.0          & 54.4   & 82.0           & 69.9    & 67.3          & 54.4   & 76.7            & 59.6    \\
                          \rowcolor{LightCyan}
                          & TSGM (Ours)              & \textbf{94.4} & \textbf{92.1} & \textbf{92.6} & \textbf{84.3} & \textbf{70.3} & \textbf{62.8} & \textbf{85.7} & \textbf{79.7}    \\ \midrule
% \multirow{4}{*}{Curved}    & RL (10M step)                & 7.5           & 3.2    & 9.5            & 7.1     & 5.5           & 4.7    & 7.5             & 5.0     \\
                        %   & RL (extra data + 100M steps) & 22.2          & 16.5   & 20.7           & 18.5    & 4.2           & 3.7    & 15.7            & 12.9    \\
                        %   & BC w/ ResNet + Metric Map    & 3.1           & 2.5    & 0.8            & 0.7     & 0.2           & 0.1    & 1.3             & 1.1     \\
                        %   & BC w/ ResNet + GRU           & 3.6           & 2.8    & 1.1            & 0.9     & 0.5           & 0.3    & 1.7             & 1.3     \\
                        %   & NRNS~\cite{NRNS}                         & 36.5          & 18.3   & 23.9           & 12.0    & 12.5          & 6.8    & 24.3            & 12.4    \\
\multirow{3}{*}{Curved}                          & NRNS~\cite{NRNS}                        & 31.7 & 13.0 & 29.0 & 13.6 & 19.2 & 10.4 & 26.6 & 12.3 \\
                          & VGM~\cite{VGM}                          & 81.0          & 45.5   & 78.8           & 59.5    & 62.2         & 46.9   & 74.0            & 50.6    \\ 
                          \rowcolor{LightCyan}
                          & TSGM (Ours)              & \textbf{93.6} & \textbf{91.0} & \textbf{89.7} & \textbf{77.8} & \textbf{64.2} & \textbf{55.0} & \textbf{82.5} & \textbf{74.1}   \\ \bottomrule
\end{tabular}}}
\label{tab:imagegoal_nrns}  
\end{table*}
\subsection{Results} 
% There are three tasks as we formulated: image goal navigation, object goal navigation, and object-context goal navigation. 

\paragraph{TSGM outperforms baselines.} 
Table~\ref{tab:imagegoal_vgm} shows the performance of our TSGM and relevant memory-based baselines on test splits of the Gibson dataset.  
Our TSGM algorithm outperforms the baseline methods in terms of Success and SPL @ 1$m$. 
TSGM improves upon the implicit memory model~\cite{TargetDriven} across splits of Gibson by 31.8\% on the success rate. 
Compared to the VGM~\cite{VGM}, our method improved by +5\% on the success rate the 4.1\% on SPL.
% and  indicating that TSGM reduced the length to the goal using the object cues.
% 
To measure performance in various situations, the test episodes are divided according to whether the path is curved or not in Table~\ref{tab:imagegoal_nrns}. 
Our TSGM shows surprising performance improvement in both curved and straight situations. 
In particular, for an easy episode, the SPL increased quite a surprising amount, 45.5\% to 91.0\% (100\%) for curved episodes and 54.4\% to 92.1\% (69.3\%) for straight episodes compared to the last state-of-the-art result of VGM~\cite{VGM}, which can be seen that if an object existing at the target position is present in the field of view, the agent can find the goal position very efficiently.
In averaged results in Table~\ref{tab:imagegoal_nrns}, TSGM outperforms competitive baselines by 8.5-9.0\% on the success rate and 20.2-23.5\% improvement on SPL.
In summary, TSGM significantly outperforms competitive baselines by 5.0-9.0\% on the success rate and 7.0-23.5\% improvement on SPL.
% 
% TSGM performs better in more challenging curved scenarios and SPL in particular increased significantly, indicating that TSGM can find efficient paths to the target.

\begin{figure}[t!]{\centering\includegraphics[width=\linewidth]{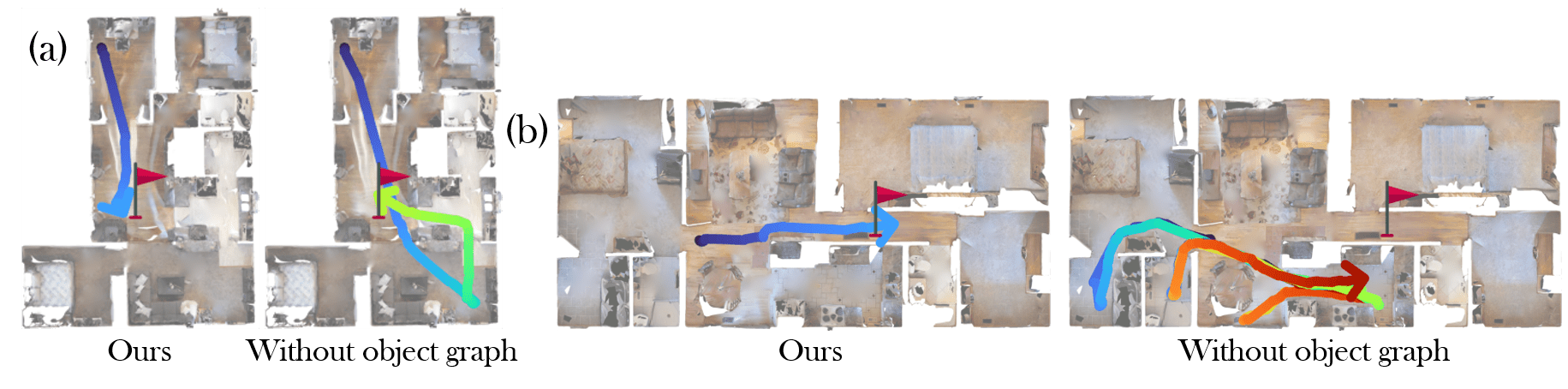}}\centering\squeezeup
\caption{\textbf{Visualization of path changes when object graph is not given.}
Our TSGM is compared with the method that does not have the object graph. The red flag highlights the image goals. The lines with an arrow represent trajectories of the agent with the color changes with time. (a) shows that an inefficient path is created when the object graph is not given. (b) The agent is trapped in the deadlock states without the object graph.
} \label{fig:traj_example}
\end{figure}

\paragraph{TSGM finds efficient paths.}
We checked that the SPL, which indicates the efficiency of the path, is improved a lot.
To investigate how the object graph assists the agent in choosing efficient paths, we visualized the paths in Figure~\ref{fig:traj_example}. 
% 
% Figure~\ref{fig:traj_example} shows how an agent finds a more efficient path when an object graph is not provided for the same episode.
% 
When an object graph is not provided, an agent passes the goal without realizing it is a goal, while an agent with an object graph quickly understands that the goal has been achieved and terminates the search (see Figure~\ref{fig:traj_example}(a)).
% 
% In other words, the object graph can be considered as assisting in the certainty of a goal through semantic knowledge and helps to find an efficient path.
% 
In Figure~\ref{fig:traj_example}(b), the agent enters a deadlock when the object graph is not provided.
In the case of a TSGM with an object graph, on the other hand, the goal is successfully found with a more efficient path.

\paragraph{Cross update module is effective.}
\begin{wraptable}{o}{0.27\textwidth}\squeezeup
% \begin{table}[h]
\centering
\caption{{Update rules.}}
{\resizebox{\linewidth}{!}{
\begin{tabular}{l|cc}\toprule
\textbf{Update} & \multicolumn{1}{c}{\textbf{Success}} & \multicolumn{1}{c}{\textbf{SPL}} \\ \midrule
No & 0.533 & 0.393 \\
Visual & 0.578 & 0.446 \\
Object & 0.613 & 0.458 \\
Cross & \textbf{0.627} & \textbf{0.471} \\  \bottomrule 
\end{tabular}}} \label{tab:ablation}
% \end{table}
\end{wraptable}
As shown in Table~\ref{tab:ablation}, the ablation study on the graph update was performed, where we only used imitation learning results on hard episodes for the ablation experiments. 
The object update version indicates the update is done only to the image node to the object node, and the visual update version shows the results when the update is the opposite.
Using the semantic graph update shows a 4.5\% improvement in success rate and +5.3\% on SPL. 
Surprisingly, the updating object node with visual node shows better results than updating graph with opposite direction, showing +8.0\% on the success rate and +6.5\% on SPL compared to the No update version. 
Finally, the cross update shows the most powerful results, +9.4\% on the success rate and +7.8\% on SPL.
The results indicate that cross updating improves performance by making image and object features informative.

\begin{figure}[t!]{\centering\includegraphics[width=\linewidth]{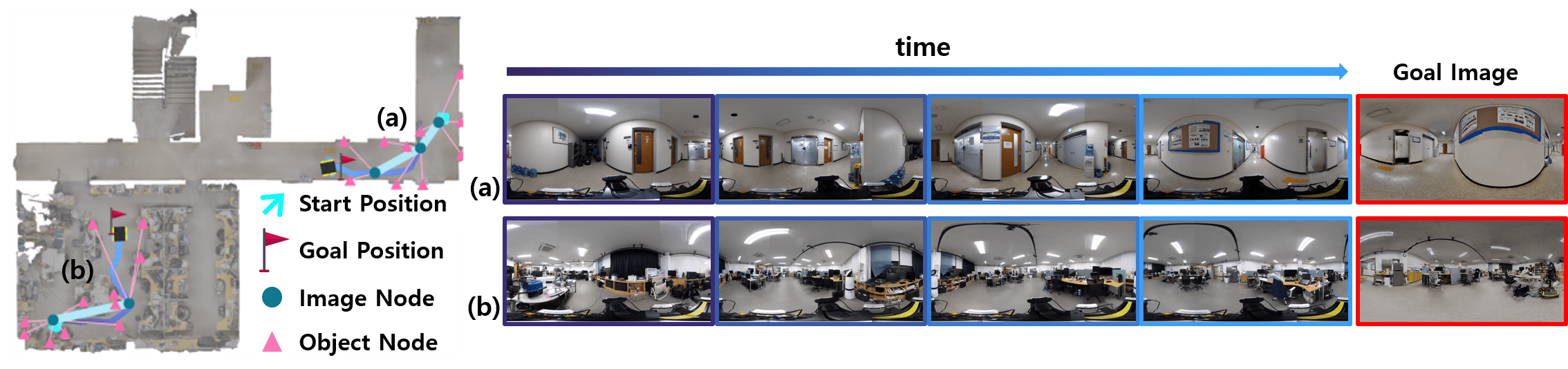}}\centering\squeezeup
\caption{\textbf{Visualization of the robot path on real environment.} The color gradation represents the flow of time. The blue arrow symbolizes the starting point, and the red flag indicates the destination. An image node is represented by a blue circle, and an object node is represented by a pink triangle.}\squeezeup \label{fig:real_robot}
\end{figure}

\paragraph{TSGM can handle the real noisy world.}
% Figure~\ref{fig:real_robot} illustrates the robot path and the constructed topological semantic graph while navigating.
Figure~\ref{fig:real_robot} shows paths and topological semantic graphs from navigation experiments using a Jackal robot.
% 
% As shown in the graph, the robot successfu
% 
The episodes examined in the real world are sampled from \textit{hard} episodes with curves.
% 
% Furthermore, since the two paths depicted are curved, they are very challenging episodes.
% 
Nonetheless, our agent demonstrated successful navigation by stopping at the correct goal location.
The supplementary video shows more paths, with the graph being incrementally built over time.

% The robot's path is visualized in Figure~\ref{fig:real_robot}.
% Since TSGM does not utilize an input for an actual position, it can determine the current position and produce an appropriate path to the goal even without a precise metric map.

 \squeezeup
% \paragraph{TSGM is robust to environmental changes.}

% \subsubsection{Object-Context Goal Navigation}
% For object-context goal navigation, we formulated an agent to find an object in an environmental context. For example, when we want to find a bag on a dining table, an agent needs to find a place that satisfies both conditions; a place with a bag and a dining table. 

\section{Conclusion and Limitation}
The proposed method explores while incrementally generating a semantic topological graph using landmarks such as objects.
The core idea of our method is derived from animal behavior, which uses landmarks as a navigational cue.
Contextual information is taken from the constructed graph to discover an efficient path to the goal.
% , which represents the current and goal location.
% 
% With contextual memory, a policy network is trained to discover an efficient path to the goal.
% 
We demonstrate that TSGM performs efficiently and effectively, with significant performance improvements, particularly in SPL.
The proposed method is demonstrated using a Jackal mobile robot to show its effectiveness for practical visual navigation in the real-world.
% Furthermore, we demonstrate our method on a jackal mobile robot for the real indoor scenarios.
% a difficult real-world setting.

While it is expected to be suitable for finding objects since object contexts are successfully trained in TSGM, a limitation of the proposed method is that no experiments on object-goal navigation are conducted{\color{black}, due to time constraint.}
We may adapt TSGM to object-goal navigation in the future.

% Although the agent seeks and navigates the goal location through estimating the position with interpolation, one limitation of our method is that it is difficult to precisely infer the goal feature for unseen locations.
% 

% 
% Therefore, the next stage in the research could be to focus on building a more general memory rather than an episodic memory in order to better estimate the goal node.
%===============================================================================
\clearpage
\acknowledgments{This work was supported by the Institute of Information \& communications Technology Planning \& Evaluation(IITP) grant funded by the Korea government(MSIT) (No. 2019-0-01309, Development of AI Technology for Guidance of a Mobile Robot to its Goal with Uncertain Maps in Indoor/Outdoor Environments)}
\bibliography{navi}  % .bib
\clearpage

\appendix
\begin{appendices}
{\color{black}

\section{Training Details and Experimental Settings}~\label{sec:sup1}
\paragraph{Training process.} 
The training process is as follows: 
1. The agent builds a graph map, which can be incrementally updated based on the observation. 2. Extract latent contextual representation from memory generated from a graph map using the attention method. 3. Extract the action by putting the attended memory, current observation, and prior action into the action policy function. 4. The action is first optimized with imitation learning with oracle action, and then the agent learns more suited for the environment through reinforcement learning.
}
    
\paragraph{Implementation details.} 
Baselines and the proposed method, excluding the methods that do not use training~\cite{NRNS}, are trained using imitation learning with 14,400 episodes, 200 episodes per scene for 72 training scenes. Then, reinforcement learning is applied for 10M frames to fix erroneous behaviors by interacting with simulation environments. {\color{black} The cross-update is carried out twice for both graphs since we wanted to update nodes using updated nodes while utilizing the least computational costs.}

\begin{wrapfigure}{o}{0.3\textwidth}
% \begin{figure}[h]
{\centering\includegraphics[width=\linewidth]{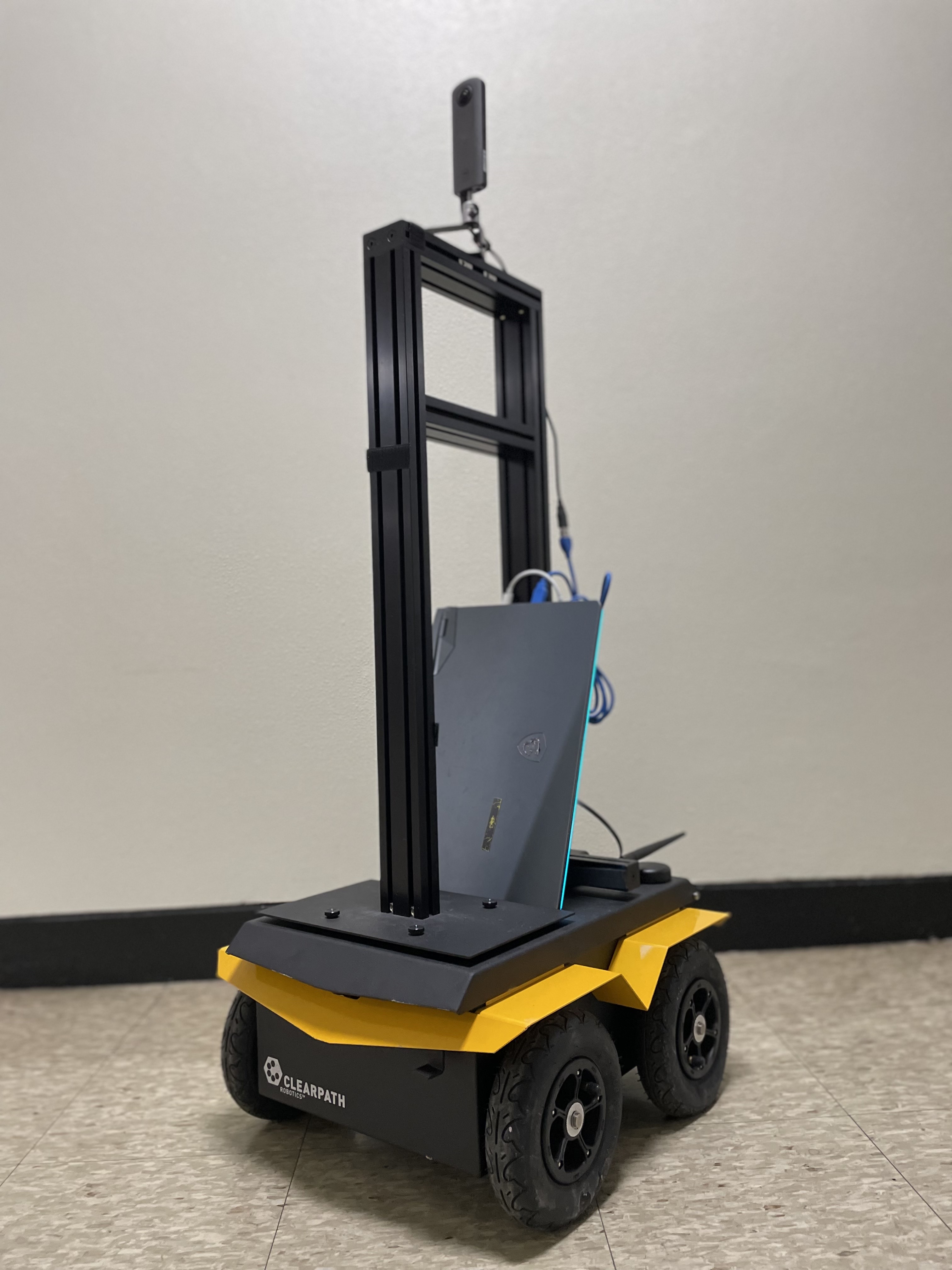}}
\caption{
Clearpath Jackal
} \label{fig:jackal}
% \end{figure}
\end{wrapfigure}

\paragraph{Real robot experiment settings.} 
We implement TSGM on a Clearpath Jackal platform for a real-world setting. 
Jackal is equipped with a single Ricoh Theta V camera, a 360$\degree$ RGB sensor.
Since this camera collects images with equirectangular projections, the equirectangular image is converted into a panoramic image of the 12 pinhole cameras having 30$\degree$ FoV which is used for training.
Since it was found that getting the exact panoramic depth is challenging, depth was extracted from the pretrained depth estimator, using the Omni-Depth~\cite{omnidepth} algorithm for the real-world experiments.  
Omni-Depth~\cite{omnidepth} estimates depth from panoramic RGB images gathered from indoor datasets such as Matterport dataset~\cite{Matterport3D}. 
We utilized the same test simulator environment for training and testing the depth estimator since we assumed a very accurate depth input was given.
To control the Jackal, we use a PD controller. 

\paragraph{Episode settings.} 
{\color{black}Gibson dataset is divided into 72 train scenes and 14 test scenes for image goal navigation. The test episodes in the Gibson dataset are derived from 14 test scenes that do not overlap the training scene.
The test episodes for Table 1 of the paper are proposed in VGM~\cite{VGM}. The number of episodes for each difficulty is 1,007, totaling 3,021 episodes.
Since episodes with similar beginning and end points might be chosen as duplicates if the scene is tiny, the number of test episodes was chosen in proportion to the scene's size.

The test episodes used in Table 2 of the paper are proposed in NRNS~\cite{NRNS}, which are divided into two types: straight and curved. The ratio of shortest path geodesic-distance to euclidean-distance between the starting and target locations in straight episodes is 1:2, and the rotational difference between the start position and destination is 45°. All other start-goal location pairs are labeled as curved episodes. 
There are 2,806 straight episodes and 3,000 curved episodes in total. 
Similar to the episode configuration used in Table 1, there are around 1,000 items divided according to the difficulty level.}

\paragraph{Similarity thresholds.}  
Two pretrained encoders are used to determine whether to add or update the graph with a new observation when building TSGM.
We adopt PCL~\cite{li2021prototypical} to obtain image similarity and Supervised contrastive learning~\cite{khosla2020supervised} d to train object similarity.
We apply the pretrained PCL~\cite{li2021prototypical} model and set the image similarity threshold ($\tau_i$) to 0.75.
We trained the model~\cite{khosla2020supervised} for object similarity utilizing provided identity values for objects on Gibson train environments as labels.
After training, the object threshold is chosen to distinguish objects with the greatest margin of error when tested on the Gibson test scene.
To evaluate the trained object encoder, we apply clustering metrics, normalized mutual information (NMI) and purity.
NMI is calculated by taking the mutual information between the true object identity distribution and the estimated distribution and then normalizing the value by the self-information of the two distributions.
Furthermore, purity is defined as the number of majority positives divided by the total number of samples.
With 0.839 NMI and 0.878 purity, the trained object encoder performed well in predicting object identifications on the clustering measures.
As a result, the object threshold ($\tau_o$) is set as 0.8 for all experiments.

\section{Graph Memory Construction}~\label{sec:sup2}

\begin{algorithm}[!h]
\DontPrintSemicolon
  \KwInput{Sequence of the input images, $\mathcal{B}$}
  \KwOutput{Topological semantic graph, $\mathcal{G}$}
  Initialize empty vertex set, $\mathcal{V}_{im}$ $\leftarrow$ $\varnothing$ , $\mathcal{V}_{ob}$ $\leftarrow$ $\varnothing$ \\
  Detect objects from the input image, $\textbf{z}_1$ $\leftarrow$ Detector($I_1$) \\
%   Encode objects, $\{o_1,...,o_K/}$ $\leftarrow$ $enc_{obj}(\textbf{z}_1)$   \\
  Initialize the last localized image, $I_l$ $\leftarrow$ $I_1$ \\ 
  Initialize the neighboring objects $O_l$ $\leftarrow$ $\textbf{z}_1$ \\ 
  Let $\mathcal{F}^i_l$ $\leftarrow$ $enc_{im}(I_l)$; $x_l$ $\leftarrow$ $\langle \mathcal{F}^i_l \rangle$ \\
  \For{\normalfont $I_t$ $\in$ $\mathcal{B}$, each observation in the batch,}{
      Encode image, $\mathcal{F}^i_t$ $\leftarrow$ $enc_{im}(I_t)$; $x_t$ $\leftarrow$ $\langle \mathcal{F}^i_t \rangle$ \\
      \If{\normalfont current place is where an agent already explored, i.e., $\exists x_m$ $\in$ $\mathcal{V}_{im}$ s.t. $cos\_dist(\mathcal{F}^i_m, \mathcal{F}^i_t)$ $\geq$ $\tau_i$ and CO($x_m$, $x_t$) $\textgreater$ 10\%,}{
            Update the most similar image node in the memory using new feature: \\
            $k$ $\leftarrow$ $\text{argmax}_m cos\_dist(\mathcal{F}^i_m, \mathcal{F}^i_t)$ \\ 
            $\mathcal{F}^i_k$ $\leftarrow$ $\mathcal{F}^i_t$ \\
            Connect the $k$th node to the last localized node: \\
            $\mathcal{E}_{im} $ $\leftarrow$ $\mathcal{E}_{im}$ $\cup$ $(x_k, x_l)$\\
        	%todo: update
        }
     \ElseIf{\normalfont $I_t$ is new, i.e., $cos\_dist(\mathcal{F}^i_l, \mathcal{F}^i_t)$ $\leq$ $\tau_i$ or $t$=0, }{
        Add $x_t$, which is $\langle \mathcal{F}^i_t \rangle$ to the graph $\mathcal{G}$: \\
        $\mathcal{V}_{im} $ $\leftarrow$ $\mathcal{V}_{im}$ $\cup$ $\{x_t\}$\\
        $\mathcal{E}_{im} $ $\leftarrow$ $\mathcal{E}_{im}$ $\cup$ $(x_t, x_l)$\\
        Update the last localized image node: \\
        $I_l$ $\leftarrow$ $I_t$\\
        Detect objects from the input image, $\textbf{z}_t$ $\leftarrow$ Detector($I_t$) \\
        Sort objects by detection scores. \\
        \For{\normalfont $\forall b_{k}$ $\in$ $\textbf{z}_t$, each object in the current observation,}{
        Encode objects, $\mathcal{F}^o_k$ $\leftarrow$ $enc_{obj}(b_k)$ \\
          \If{\normalfont $b_k$ exists in the neighboring objects $O_l$, i.e., $\exists z_m$ $\in$ $O_l$ s.t. $cos\_dist(\mathcal{F}^o_m, \mathcal{F}^o_k)$ $\geq$ $\tau_o$ and $c_k$ = $c_m$ and $r_k$ $\textgreater$ $r_m$ }{
            Update the most similar object node in the $O_l$ using new feature: \\
            $k$ $\leftarrow$ $\text{argmax}_m cos\_dist(\mathcal{F}^o_m, \mathcal{F}^o_t)$ \\ 
            $\mathcal{F}^o_k$ $\leftarrow$ $\mathcal{F}^o_t$
          }
          \ElseIf{\normalfont Input object is new according to the neighboring objects, i.e., $cos\_dist(\mathcal{F}^o_m, \mathcal{F}^o_k)$ $\leq$ $\tau_o$, $\forall b_m$ $\in$ $\hat{\textbf{b}}_n$ or $c_k$ $\neq$ $c_m$, }{
          Add the object state $z_k$, which is $\langle \mathcal{F}^o_k, c_k, r_k \rangle$ to the graph $\mathcal{G}$: \\
          $\mathcal{V}_{ob} $ $\leftarrow$ $\mathcal{V}_{ob}$ $\cup$ $\{z_k\}$\\
          $\mathcal{E}_{c} $ $\leftarrow$ $\mathcal{E}_{c}$ $\cup$ $(x_l, z_k)$\\
          }%EndIf
         }%EndFor
          $O_l$ $\leftarrow$ $\langle \textbf{z}_t$, Neighbor($\textbf{z}_t$)$\rangle$ \\
        }%EndIf
    }
    $\mathcal{G} \leftarrow \{\mathcal{V}_{im} , \mathcal{V}_{ob}, \mathcal{E}_{im}, \mathcal{E}_{c}\}$\\
  \Return{$\mathcal{G}$}
\caption{Build Topological Semantic Graph Memory}
\label{alg:tsgm}
\end{algorithm}

When an agent is placed in an unknown environment, the agent explores the environment to acquire a graph memory. The graph memory is constructed using two pretrained encoders: an image encoder $enc_{im}$ and object encoder $enc_{obj}$.
To improve the graph with common objects, we add a rule that prevents adding a new image node if the ratio of the number of common objects of the two image nodes is larger than 10\%,
\begin{align}
\begin{aligned}
  \text{CO}(x_m, x_t) = \frac{\#(O(x_m) \cap O(x_t))}{\#(O(x_m) \cup O(x_t))} > 10\%,
\end{aligned}
\end{align}
where $O(x_m)$ represents the objects connected to the image node $x_m$.
The detailed algorithm for building a topological semantic graph is described in Algorithm~\ref{alg:tsgm}.
\paragraph{Object node.} 
Object features are retrieved using ROI-align~\cite{he2017mask} based on the position of the object bounding box.
The object features are extracted from the output of the conv4 layer of ResNet18.
% 
% The object node is composed of object features extracted from the image using ROI align~\cite{he2017mask} and the detected category of the object.
% 
The object category from the detector is encoded with two-layered neural networks and then concatenated to the object features to make an object node state $z_k = \langle \mathcal{F}^o_k, c_k, r_k \rangle$ for $k$th object. $\mathcal{F}^o_k$ is the extracted object feature, $c_k$ represents the category of the object, and $r_k$ is the detection score of the object.
The number of categories is 80 since we employ a detector that is pretrained on the COCO dataset.
\paragraph{Object node updating rules.}
The graph builder compares the object's detection scores when the same object node is found. 
If a new input object's feature is similar to an object in memory and the category is the same, the graph builder determines that the two objects are identical.
The graph builder compares the detection scores to increase the detected features' quality.
Only when the newly discovered object has a higher score, the graph builder updates the node feature.
% 

% \paragraph{Image node.} 

% \paragraph{Image node updating rules.}

\paragraph{Graph connections.}
While building the graph, the affinity matrices can be calculated based on the edge information,
\begin{align}
\begin{aligned}
  A_{im}[x_v,x_w] &= 1 \; \forall(x_v,x_w) \in \mathcal{E}_{im}, \\
  A_{c}[x_v,z_k] &= 1 \; \forall(x_v,z_k) \in \mathcal{E}_{c}. \\
\end{aligned}
\end{align}
To make objects in proximity connected in the graph memory, we calculate the affinity matrix for object nodes to be connected to objects in the neighbors of the current image node using the image affinity $A_{im}$ and image-object affinity.
The object graph affinity matrix can be calculated using the above matrices,
\begin{align}
\begin{aligned}
  A_{ob} = A_{c}^T  (A_{im} + \text{I}) A_{c},
\end{aligned}
\end{align}
where I is an identity matrix, which connects the object nodes that share the same image node.

\section{Cross Graph Mixer}~\label{sec:sup3}
The cross graph mixer module runs for $L$ steps and is defined in terms of a message function $M$ and vertex update function $U$. The message function can be seen as a composition of two functions, $M = C \circ S$, where $S$ is the self-update function, and $C$ is the cross-update function.
At first, the image memory state $x$ and object memory state $z$ are the inputs of the network,
\begin{align}
\begin{aligned}
    {hi}_v^{1} = x_v, \forall v \in \mathcal{V}_{im}\\
    {ho}_k^{1} = z_k, \forall k \in \mathcal{V}_{ob}.
\end{aligned}
\end{align}
For $l$th step, where $l \in \{1, ..., L\}$, image and object node states are self-updated to get contextual representations between nearby locations or objects,
\begin{align}
\begin{aligned}
    \hat{mi}_v^{l} = \sum_{w \in \mathcal{N}_i(v)} S_{i}^l({hi}_w^l, A_{im}, g), \\
    \hat{mo}_v^{l} = \sum_{k \in \mathcal{N}_o(v)} S_{o}^l({ho}_k^l, A_{ob}, g),
\end{aligned}
\end{align}
where $\mathcal{N}_i(v)$ and $\mathcal{N}_o(v)$ denote the image/object neighbors of the $v$th node, respectively. 
The self-update message function $S$ makes a $d$ dimensional vector by aggregating connected nodes, i.e., 
\begin{align}
\begin{aligned}
S_i^l({hi}_w, A_{im}, g) &= \text{A}_{im}[v,w]\text{B}^l_{i}(e_{vw}) f_i({hi}_w^l || g)\\
S_o^l({ho}_k, A_{ob}, g) &= \text{A}_{ob}[v,k]\text{B}^l_{o}(e_{vk}) f_o({ho}_k^l || g),
\end{aligned}
\end{align}
where $||$ is a concatenation operation and $f_i$ and $f_o$ are two-layered neural networks. Here, $\text{B}(e_{vw})$ is a learned connection relationship between $v$th and $w$th nodes, i.e., the edge vectors between the nodes, which maps the edge vector $e_{vw}$ to a $d_i \times d_i$ matrix, 
\begin{align}
\begin{aligned}
\text{B}^l_i(e_{vw}) &= \textbf{v}_i^T \text{concat}(\textbf{W}_i hi_v^l,\textbf{W}_i hi_v^l)\\ \text{B}^l_o(e_{vk}) &= \textbf{v}_o^T \text{concat}(\textbf{W}_i ho_v^l,\textbf{W}_o ho_k^l), 
\end{aligned}
\end{align}
where $\textbf{W}_{i} \in \mathbb{R}^{C \times d_i}$, $\textbf{v}_{i} \in \mathbb{R}^{2d_i \times d_i}$, $\textbf{W}_{o} \in \mathbb{R}^{C \times d_o}$,  $\textbf{v}_{o} \in \mathbb{R}^{2d_o \times d_o \times d_o}$ are the matrix parameters, which are applied for all latent nodes. 

Then, the nodes aggregate the different types of nodes during the message passing phase to make a complete message ${mi}_v^{l}$ and ${mo}_v^{l}$.
For this, nodes  cross update is done for image nodes to aggregate messages from object nodes, while object nodes use image nodes to create messages,
\begin{align}
\begin{aligned}
    {mi}_v^{l} = \sum_{k \in \mathcal{N}_o(v)} C_i^l(\hat{mi}_v^l, \hat{mo}_k^l, A_{c}), \\
    {mo}_v^{l} = \sum_{w \in \mathcal{N}_i(v)} C_o^l(\hat{mo}_v^l, \hat{mi}_w^l, A_{c}), 
\end{aligned}
\end{align}
where
\begin{align}
\begin{aligned}
C_i^l(\hat{mi}_v^l, \hat{mo}_k^l, A_c) &= \hat{mi}^l_v + f'_{i}(\text{A}_{c}[v,k]\text{B}^l_{i}(e_{vk}) f^{''}_i(\hat{mo}_k^l || g)),\\
C_o^l(\hat{mo}_v^l, \hat{mi}_w^l, A_c) &= \hat{mo}^l_v + f'_{o}(\text{A}_{c}[w,v]\text{B}^l_{o}(e_{wv}) f^{''}_o(\hat{mi}_w^l || g)),
\end{aligned}
\end{align}
where $f'_o$, $f'_{i}, f^{''}_o$ and $f^{''}_{i}$ are two-layered neural networks. 

Finally, the update function $U$ transfers messages from object nodes to image nodes, 
\begin{align}
\begin{aligned}
    {hi}_v^{l+1} = U_i^l({hi}_v^l, {mo}_v^{l}),\\
    {ho}_v^{l+1} = U_o^l({ho}_v^l, {mi}_v^{l}),
\end{aligned}
\end{align}
where,
\begin{align}
\begin{aligned}
U_i^l({hi}_v^l, mo_{v}^{l+1}) &= hi_v^{l} + g_i(A_c[\cdot,v] mo_v^{l+1}), \\
U_o^l({ho}_v^l, mi_{v}^{l+1}) &= ho_v^{l} + g_o(A_c[v,\cdot] mi_v^{l+1}),
\end{aligned}
\end{align}
where $g_i$ is a neural network that maps hidden states of an object to hidden states of the connected images and $g_o$ is a neural network that maps hidden states of an image to hidden states of the connected objects.
% and $\hat{A}_c$ is a column-wise normalized matrix of ${A}_c$.

After $L$ iterations, the output of the network become a contextual memory. The output of the image stage is,
\begin{align}
\begin{aligned}
\textbf{mi}_v &= hi^L_v \; \forall v \in \{1, ..., N\}, \\
\textbf{mo}_k &= ho^L_k \; \forall k \in \{1, ..., M\}.
\end{aligned}
\end{align}
Note that $\textbf{mi} = \{\textbf{mi}_1, ..., \textbf{mi}_N\}$ is the contextual image memory and $\textbf{mo} = \{\textbf{mo}_1, ..., \textbf{mo}_M\}$ is the contextual object memory. 
% Since each node is calculated using the cross graph mixer module, the number of input and output nodes remains constant.

{\color{black}
\section{Object-Based Methods}~\label{sec:sup4}
We compared TSGM to navigational methods~\cite{VTNET, OMT} that employ object information.
In VTNet~\cite{VTNET}, an image and objects observed by an agent are combined to produce a fused representation of the place. VTNet~\cite{VTNET} does not have any explicit memory and only has an implicit RNN memory.
TSGM, on the other hand, can explicitly employ previous knowledge gathered while navigating the environment.
VTNet~\cite{VTNET} connects object information when they are detected in the same image.
Since it combines objects detected in the same image, VTNet~\cite{VTNET} can be seen as a method that only connects objects detected in the same image. 
On the other hand, TSGM can connect neighboring object nodes even if the objects are not detected in the same image, resulting in contextually solid representations.

In Object memory transformer~\cite{OMT}, image and object representation are saved in the explicit memory every time step.
Since TSGM only puts a new node into a graph memory based on the similarity between memory and current observations for both image and object graphs, it has less redundancy than \cite{OMT}.
\cite{OMT} utilizes only the preceding $T$ chunks of data are utilized.
TSGM, on the other hand, makes use of all graph memory information derived from past exploration of the environment. 
This is achievable due to TSGM's low redundancy.
}

\begin{figure}[!t]{\centering\includesvg[width=0.6\linewidth]{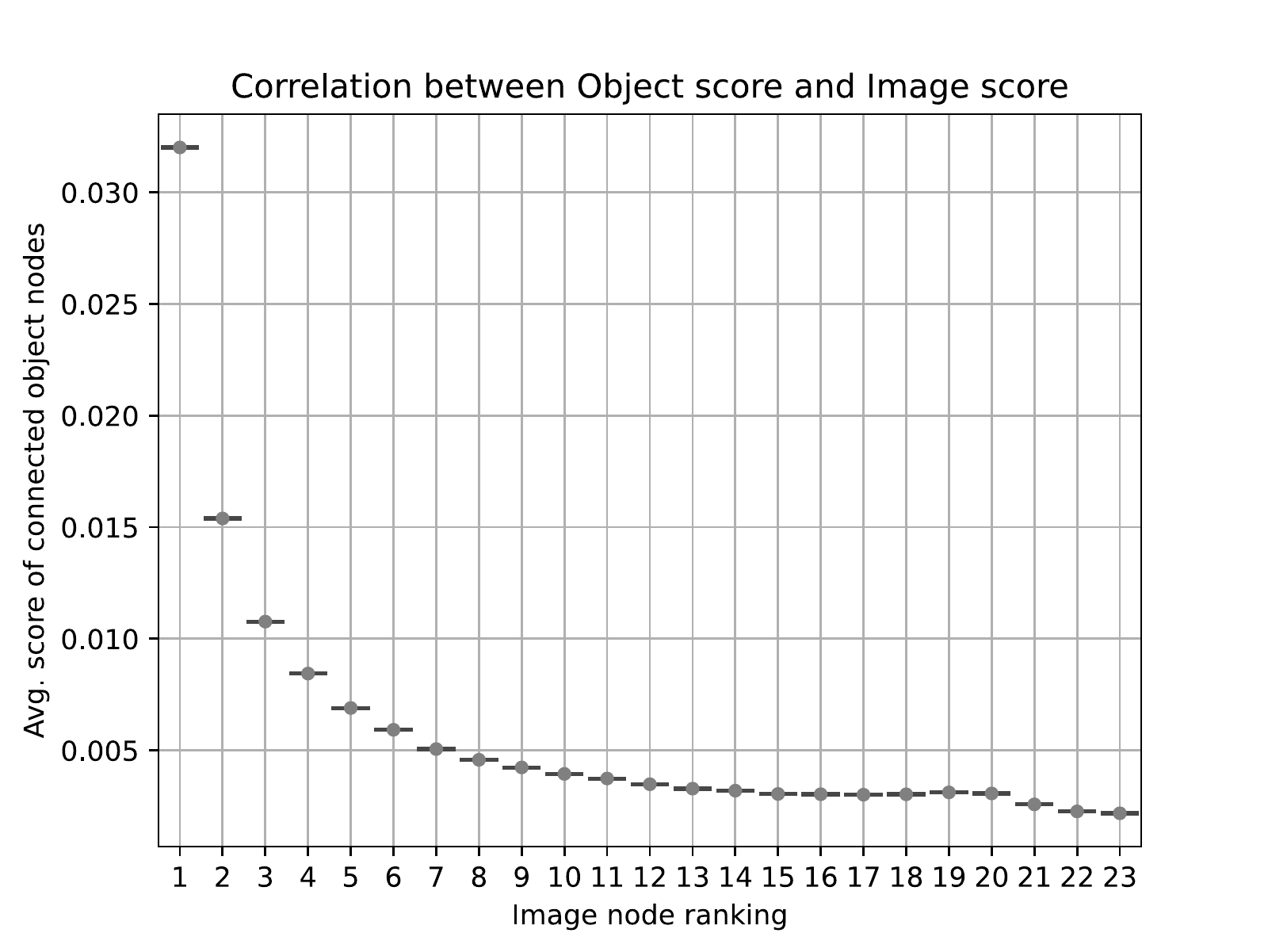}}\centering
\caption{\color{black}Correlation between the object score and image score. It indicates that the object nodes help image nodes to localize to the current node.}
\label{fig:correlation}
\end{figure}

\begin{figure}[!t]{\centering\includegraphics[width=\linewidth]{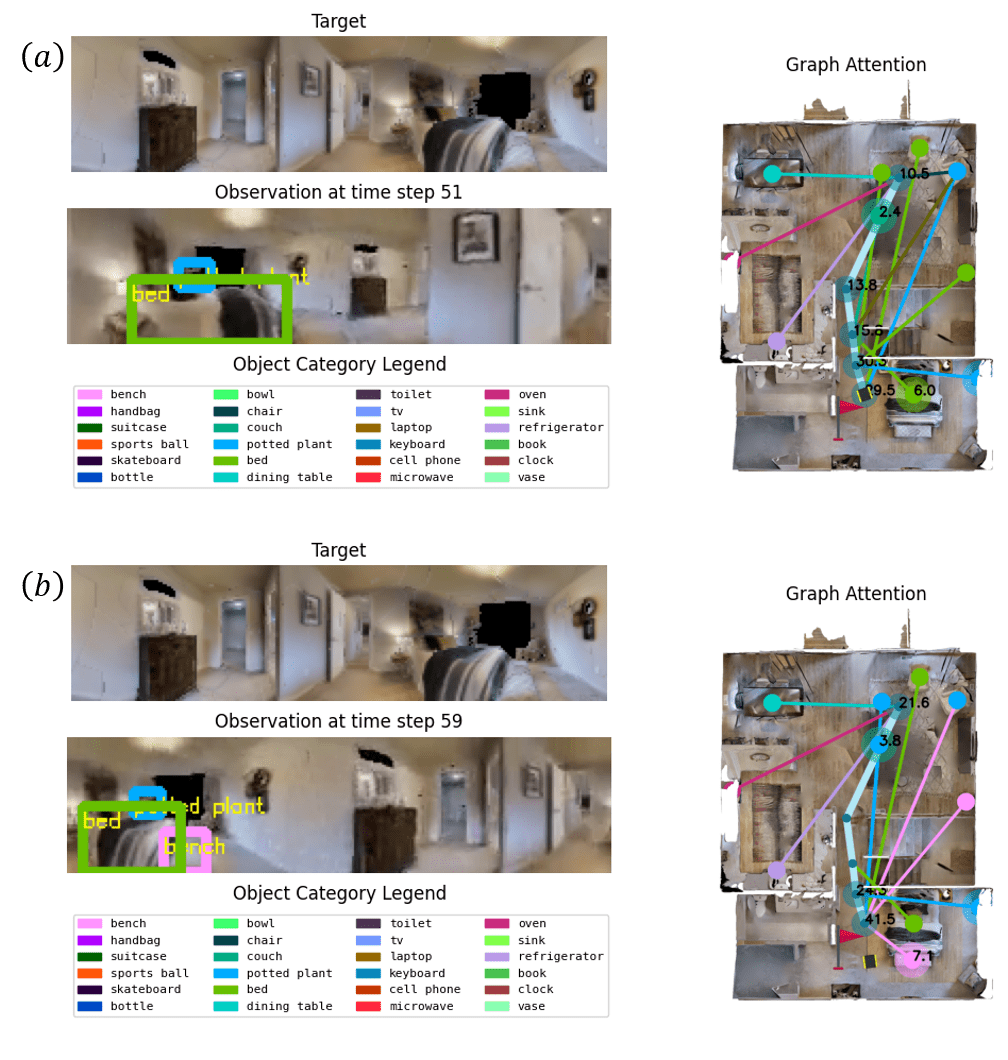}}\centering
\caption{Attention scores of the image and object nodes.}
\label{fig:attention1}
\end{figure}
\clearpage

\begin{figure}[!t]{\centering\includegraphics[width=\linewidth]{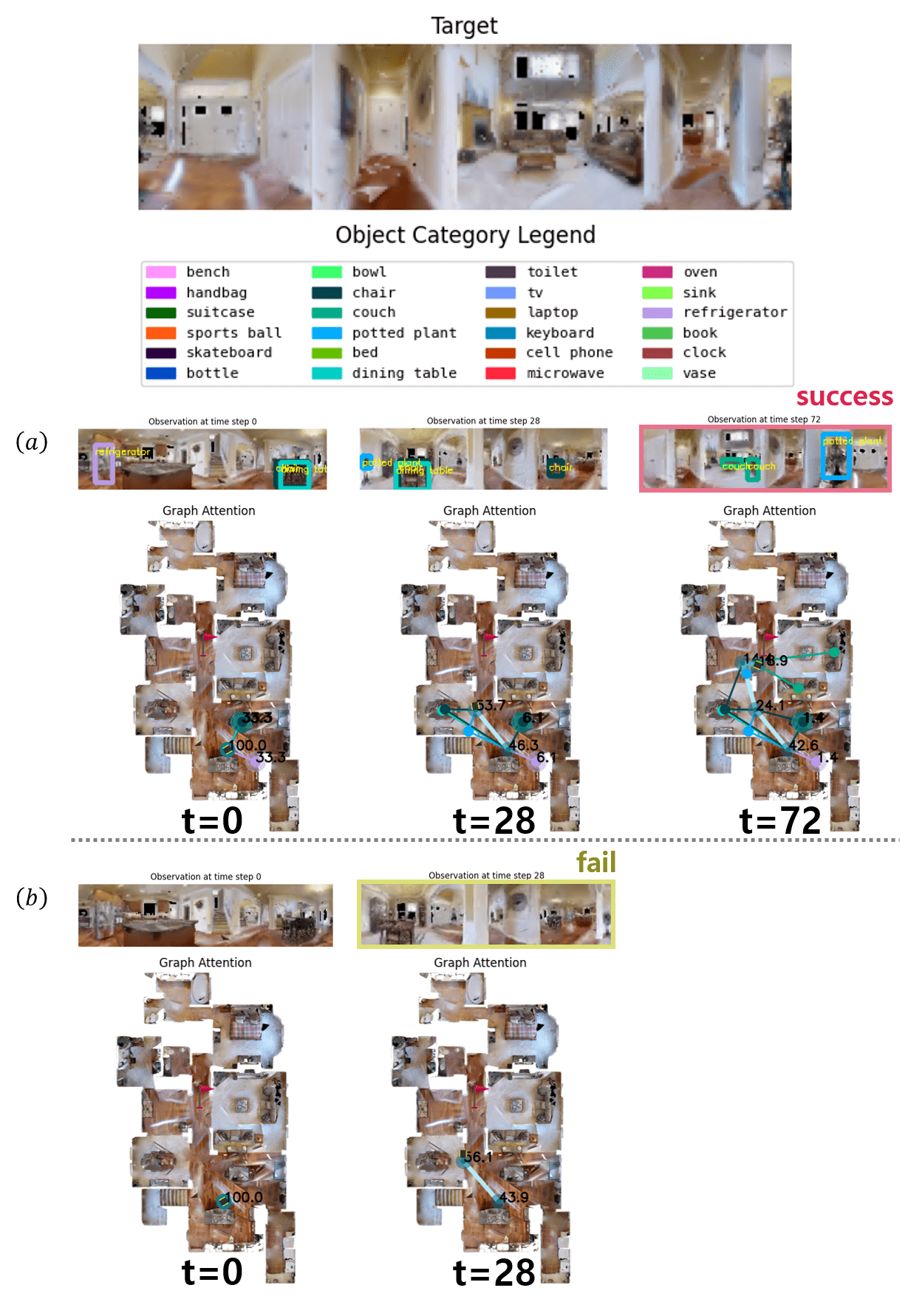}}\centering
\caption{Comparison with the method without objects. The stop button can be pressed more precisely through the object context.}
\label{fig:attention2}
\end{figure}
\clearpage

\begin{figure}[!t]{\centering\includegraphics[width=\linewidth]{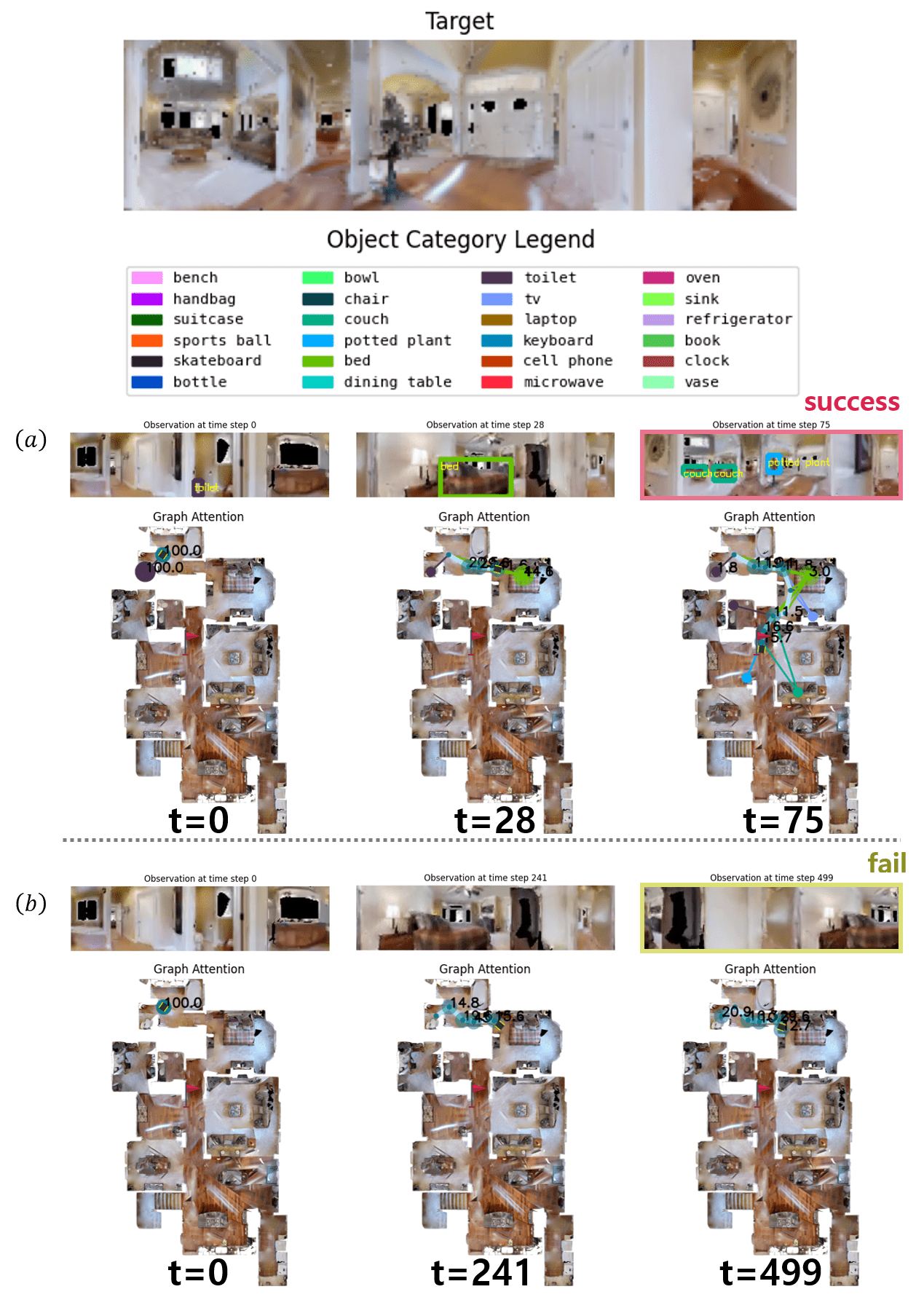}}\centering
\caption{Comparison with the method without objects. The agent can utilize object configurations to get the semantic knowledge how to go from the bathroom to living room.}
\label{fig:attention3}
\end{figure}
\clearpage

{\color{black}
\section{Impact of Landmarks}~\label{sec:sup5}

\paragraph{Object helps localization.} 
To demonstrate the influence of object information on performance improvement, we conducted additional experiments using attention values from the memory attention module (Section~\ref{sec:pap3.4}). 
The agent draws the best image memory in the memory attention module given the current observation.
We assume that localization is successful if the selected image node, which has the highest attention score, is the closest to the agent location.
The success rate of localization is computed using the 1,007 hard episodes, which are used in Table 1 of the manuscript. 
As a result, localization performance is 38.5\% without an object node whereas it is 68.2\% with an object node, an improvement of 29.7\%.

\paragraph{Object and image are correlated.} 
To demonstrate that incorporating objects assists in localization, we investigate the correlation between image and object nodes.
We averaged the obtained attention values across the objects in an image since it contains many objects.
The vertical axis in Figure~\ref{fig:correlation} reflects the average attention score of the object nodes connected with the corresponding ranking's image node.
Since the attention score tends to diminish as the number of image nodes rises, the image node ranking was employed as the horizontal axis by sorting in descending order using the attention value.
As a consequence, the image node selected as the current node demonstrated a substantial association with a higher object node score (Figure~\ref{fig:correlation}).
The likelihood of being in the adjacent image node increases as the object node score increases, suggesting that the inclusion of object nodes benefited with agent localization.
%
% To summarize, using object nodes allows the agent to search the path more efficiently since an agent is less likely to getting lost by properly localizing the agent.
As a consequence, the image node selected as the current node demonstrated a substantial association with a higher object node score.
In summary, it is claimed that the utilization of object nodes enables the agent to search the path more efficiently while avoiding getting lost by properly localizing the agent.

\paragraph{Visualization of the attention score.} 
We showed the attention scores of image and object nodes to demonstrate the significance of object features.
To use a ground truth detector, we finetuned TSGM using PPO for 2M frames in the Gibson tiny dataset for this experiment.
For the comparison experiments, a model without objects was also finetuned in the same setting, using PPO for 2M frames. 
The categories in the Gibson tiny dataset are the same as those in the COCO dataset; we displayed 24 of them.
Besides the illustration, we show the legend that indicates the color of each object category.
The image node is blue, whereas the object node is colored differently for each category. 
The attention values are multiplied by 100 and scaled to a score ranging from 0 to 100 for the best view.
In Figure~\ref{fig:attention1}, the attention scores are depicted in Figure~\ref{fig:attention1}(a) at step 51 and Figure~\ref{fig:attention1}(b) at step 59. When the \textit{bench} node is not visible, the nearest image node has a score of 29.5. 
After the \textit{bench} node is recognized, the current node's attention score climbs to 41.5. 
The score of the currently located \textit{bench} and \textit{potted plant} is high, indicating that the current image node's localization accuracy has improved.
A model in Figure~\ref{fig:attention2}(a) is trained using an object, whereas a model in Figure~\ref{fig:attention2}(b) was trained without an object.
The initial paths taken by Figure~\ref{fig:attention2}(a) and Figure~\ref{fig:attention2}(b) are similar, but Figure~\ref{fig:attention2}(b) failed since it pressed the stop button at the incorrect place. 
The model in Figure~\ref{fig:attention2}(b) was unsuccessful since it could not localize at the goal. 
On the other hand, Figure~\ref{fig:attention2}(a) successfully hit the stop button precisely at the target location after realizing that the place at time step 28 was not the target position by using the object configuration information.
Figure~\ref{fig:attention3} shows the result of the model trained without an object (Figure~\ref{fig:attention3}(b)) and the version trained with an object (Figure~\ref{fig:attention3}(a)). 
The goal is successfully reached by Figure~\ref{fig:attention3}(a) in 75 steps, but Figure~\ref{fig:attention3}(b) demonstrates that it does not exit the bathroom. 
This issue appears since the agent failed to acquire the semantic relevant knowledge to go from the bathroom or room to the living room.}

\section{Path and Graph Visualizations}~\label{sec:sup6}
\begin{figure}[h]{\centering\includegraphics[width=\linewidth]{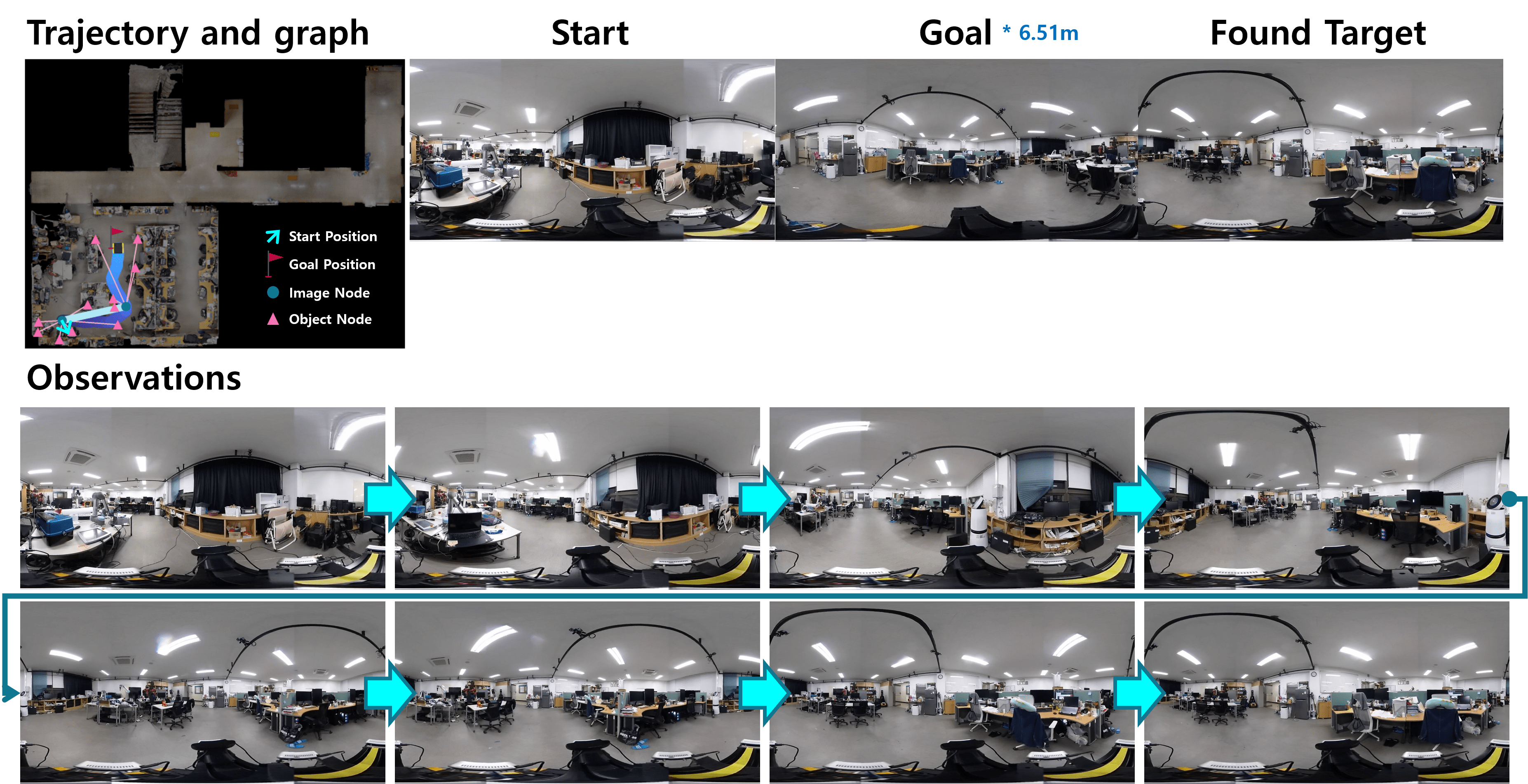}}\centering
\caption{Real robot experimental results from the laboratory environment.}
\label{fig:realexp1}
\end{figure}

\begin{figure}[h]{\centering\includegraphics[width=\linewidth]{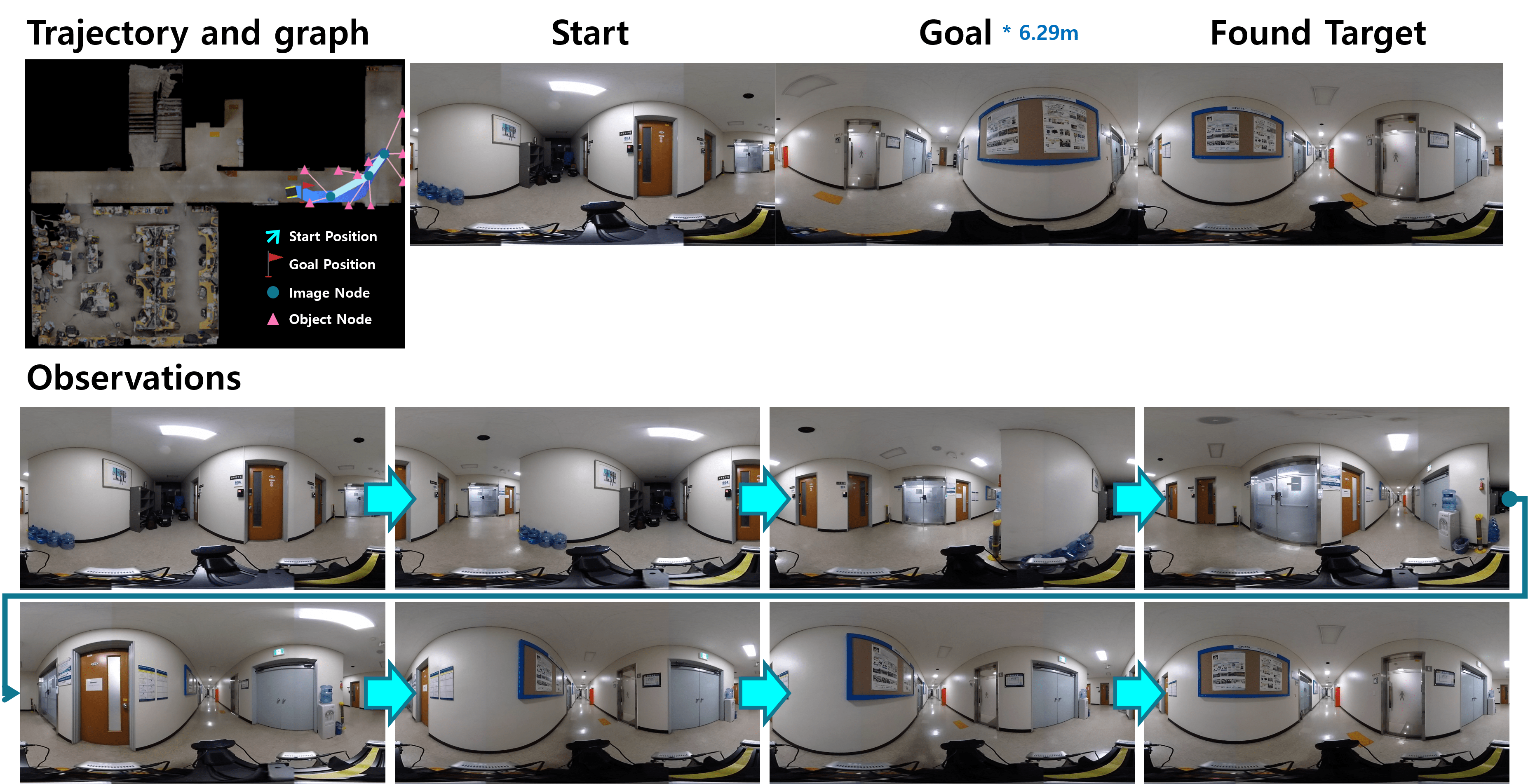}}\centering
\caption{Real robot experimental results from the laboratory environment.}
\label{fig:realexp2}
\end{figure}

\begin{figure}[h]{\centering\includegraphics[width=\linewidth]{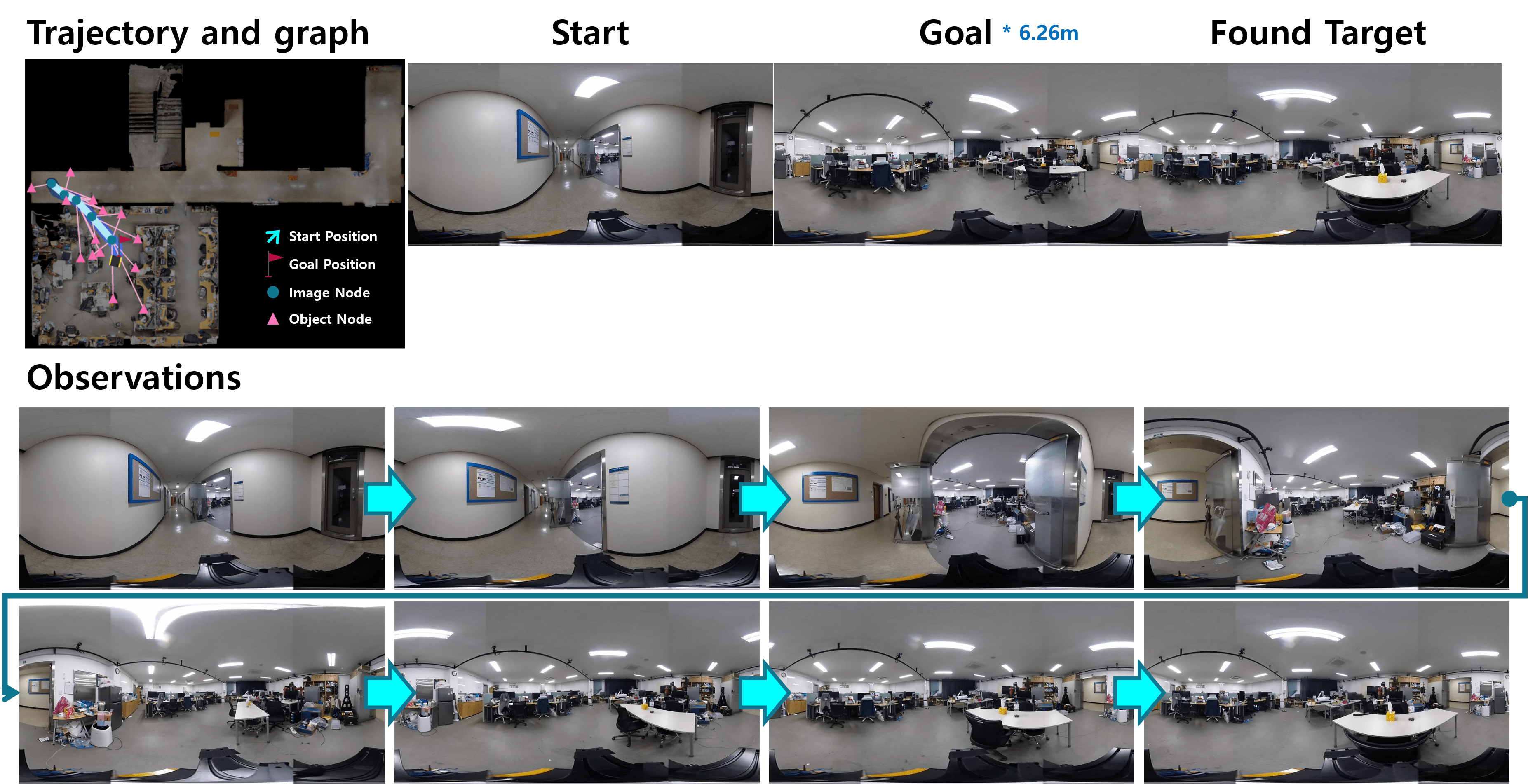}}\centering
\caption{Real robot experimental results from the laboratory environment.}
\label{fig:realexp3}
\end{figure}

\begin{figure}[h]{\centering\includegraphics[width=\linewidth]{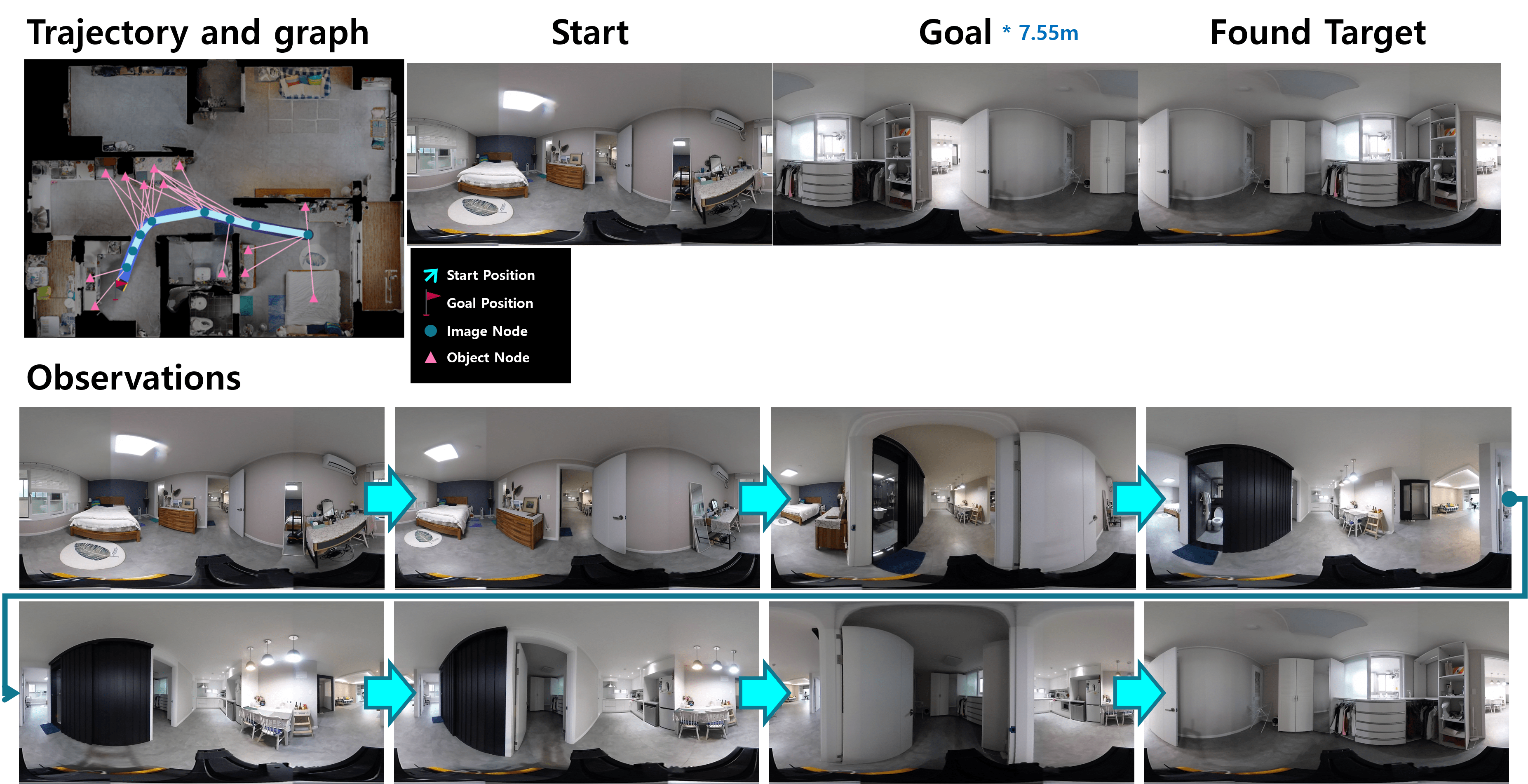}}\centering
\caption{Real robot experimental results from the home environment.}
\label{fig:realexp4}
\end{figure}

The experimental results on real world are visualized in Figure~\ref{fig:realexp1},\ref{fig:realexp2},\ref{fig:realexp3},\ref{fig:realexp4}.
All episode goals are sampled from 6$m$ to 8$m$.
Since the locations of nodes are not accurate in real world, we collected the node locations for the best view.

\begin{figure}[h]{\centering\includegraphics[width=\linewidth]{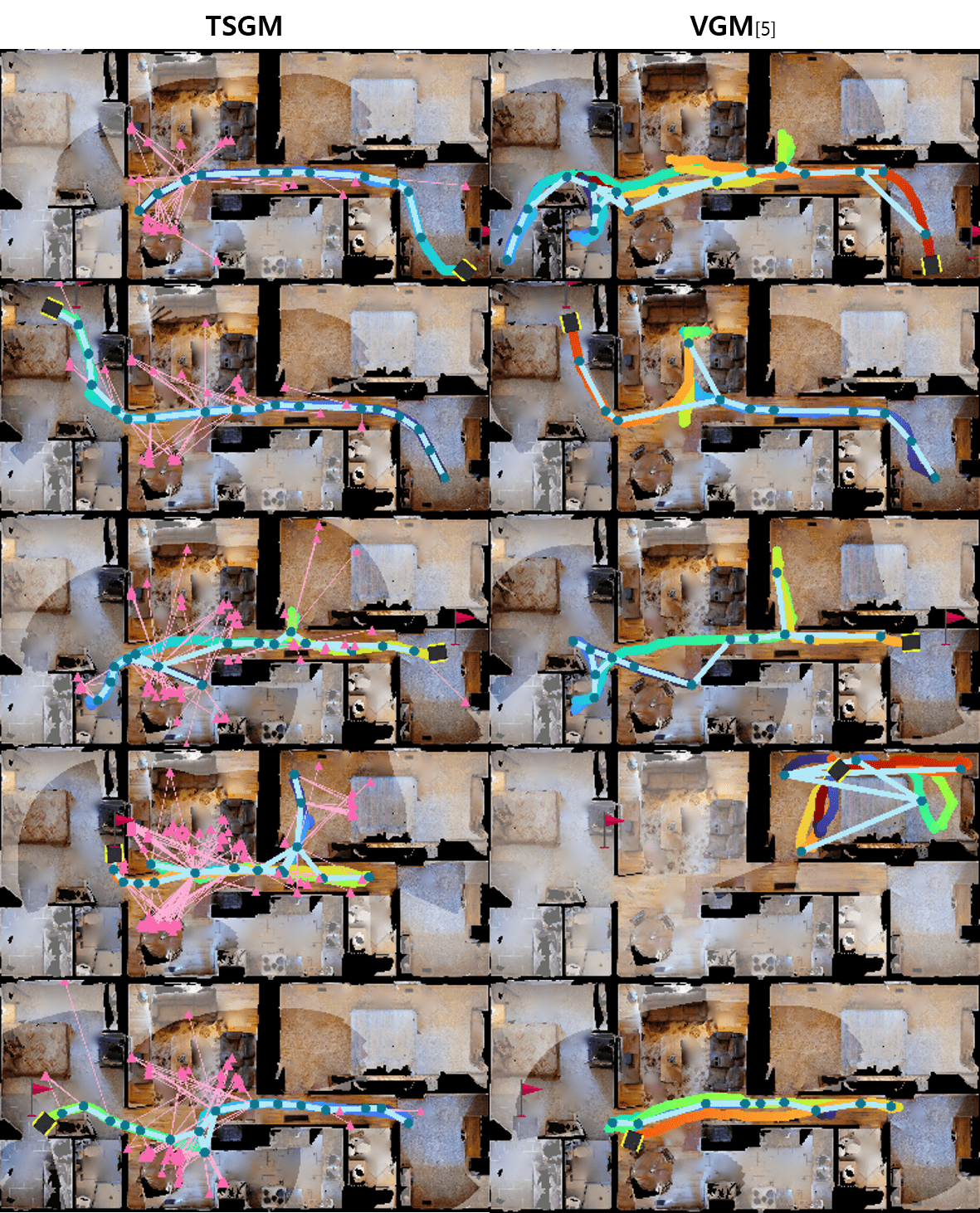}}\centering
\caption{Examples from Gibson's Cantwell environment.}
\label{fig:example1}
\end{figure}

\begin{figure}[h]{\centering\includegraphics[width=\linewidth]{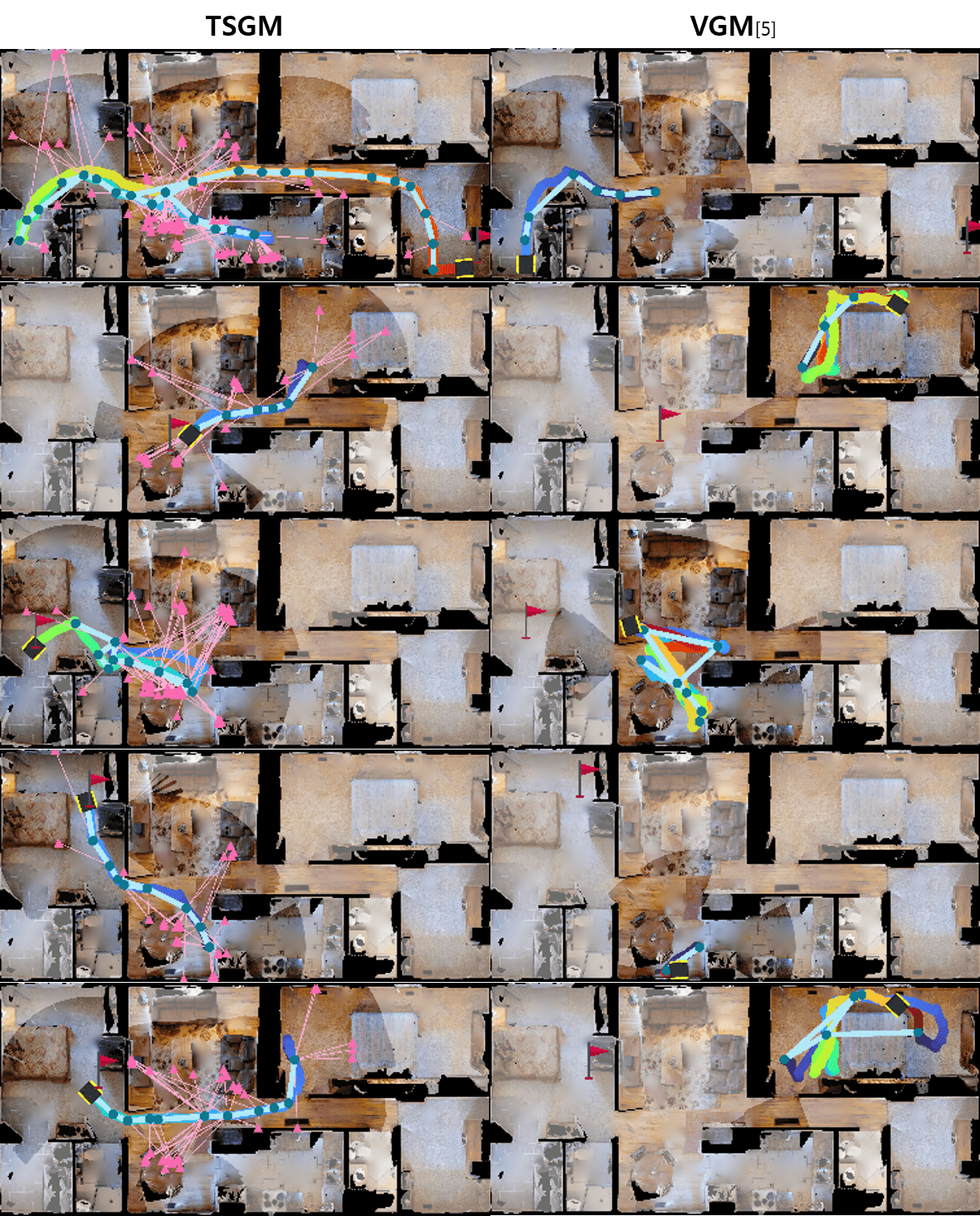}}\centering
\caption{Examples from Gibson's Cantwell environment.}
\label{fig:example2}
\end{figure}

\begin{figure}[h]{\centering\includegraphics[width=\linewidth]{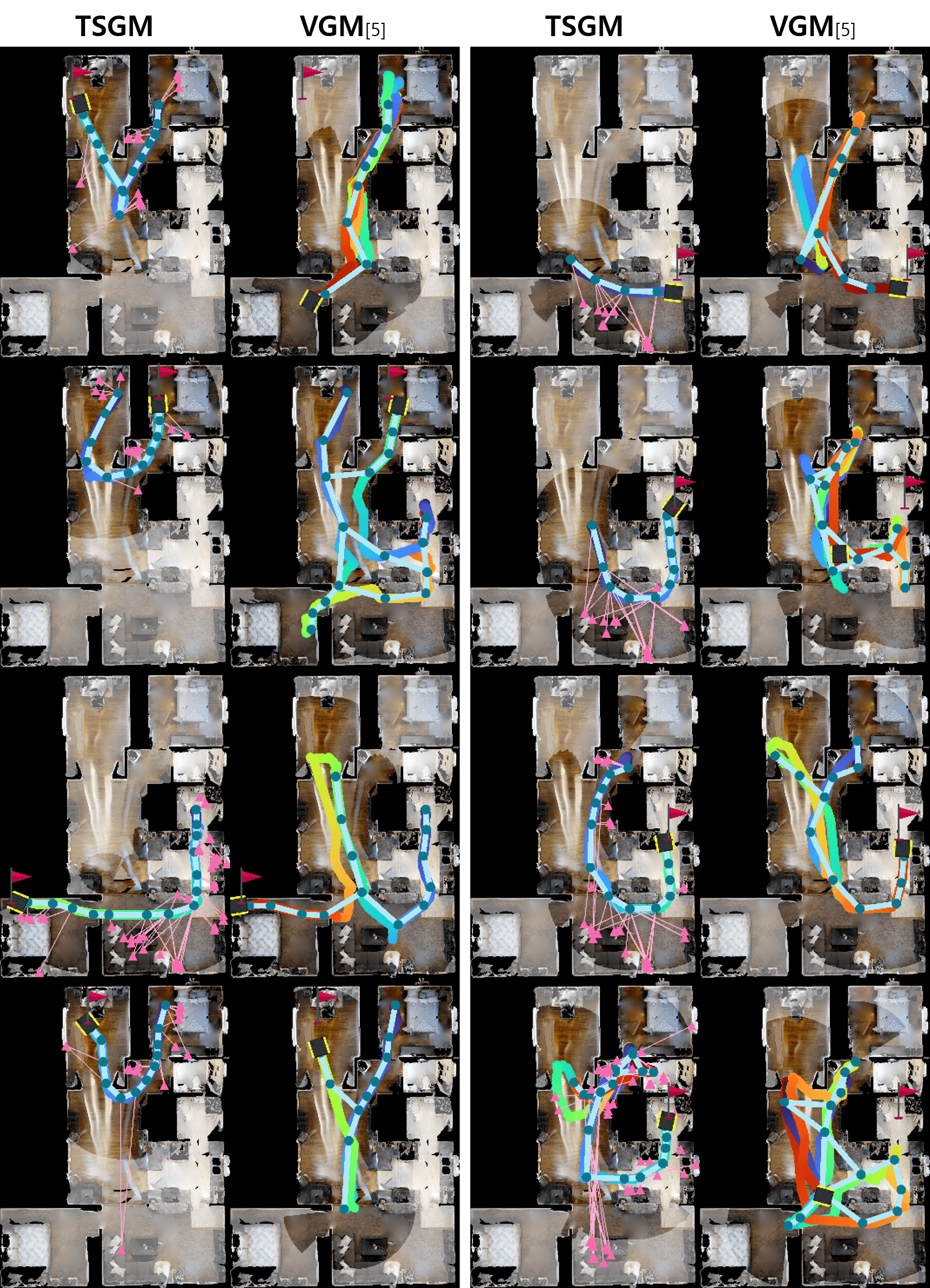}}\centering
\caption{Examples from Gibson's Eastville environment.}
\label{fig:example4}
\end{figure}

\begin{figure}[h]{\centering\includegraphics[width=\linewidth]{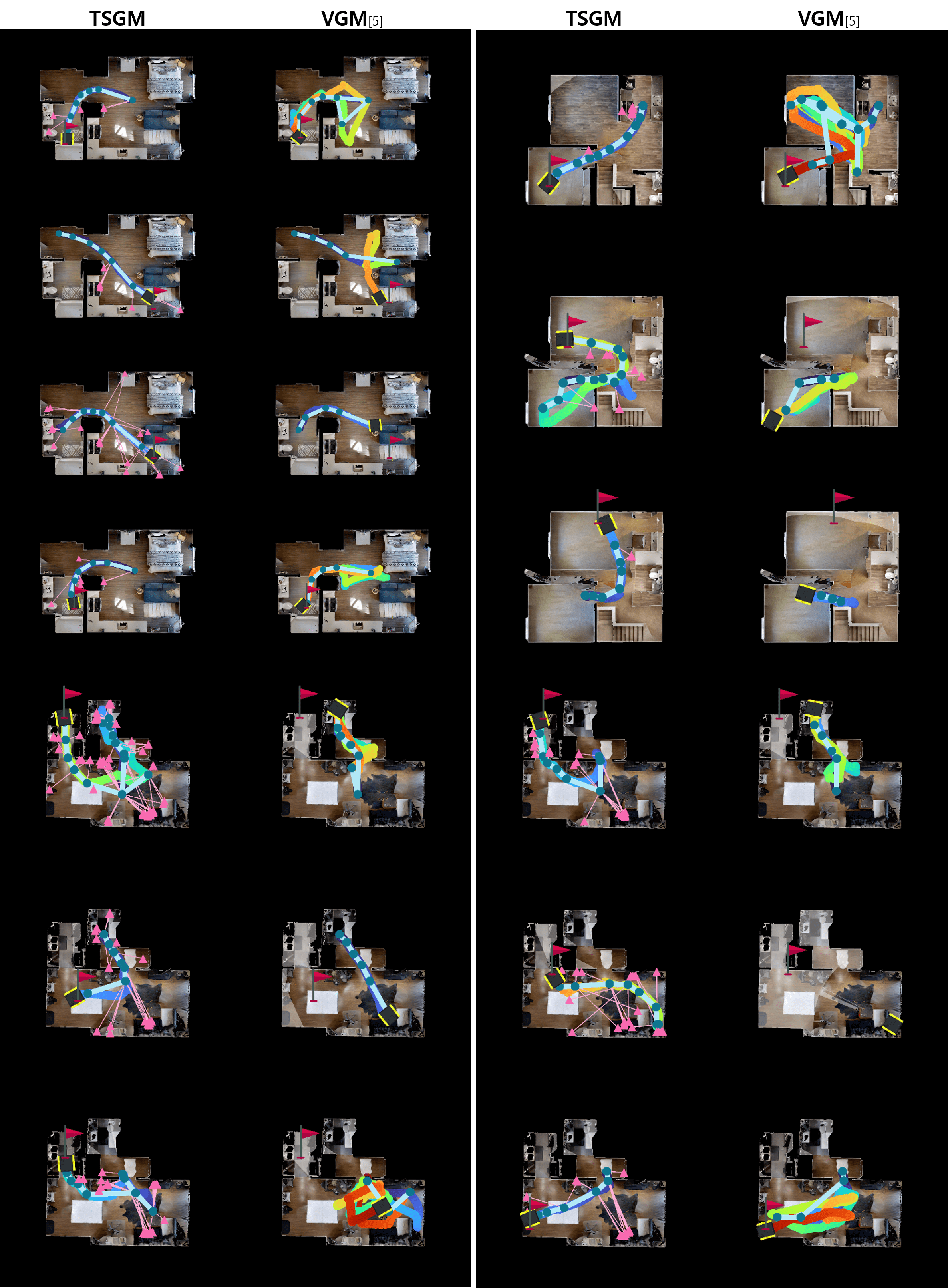}}\centering
\caption{Examples from Gibson's Denmark (bottom), Greigville (top left), Ribera (top right) environment.}
\label{fig:example5}
\end{figure}

\begin{figure}[h]{\centering\includegraphics[width=\linewidth]{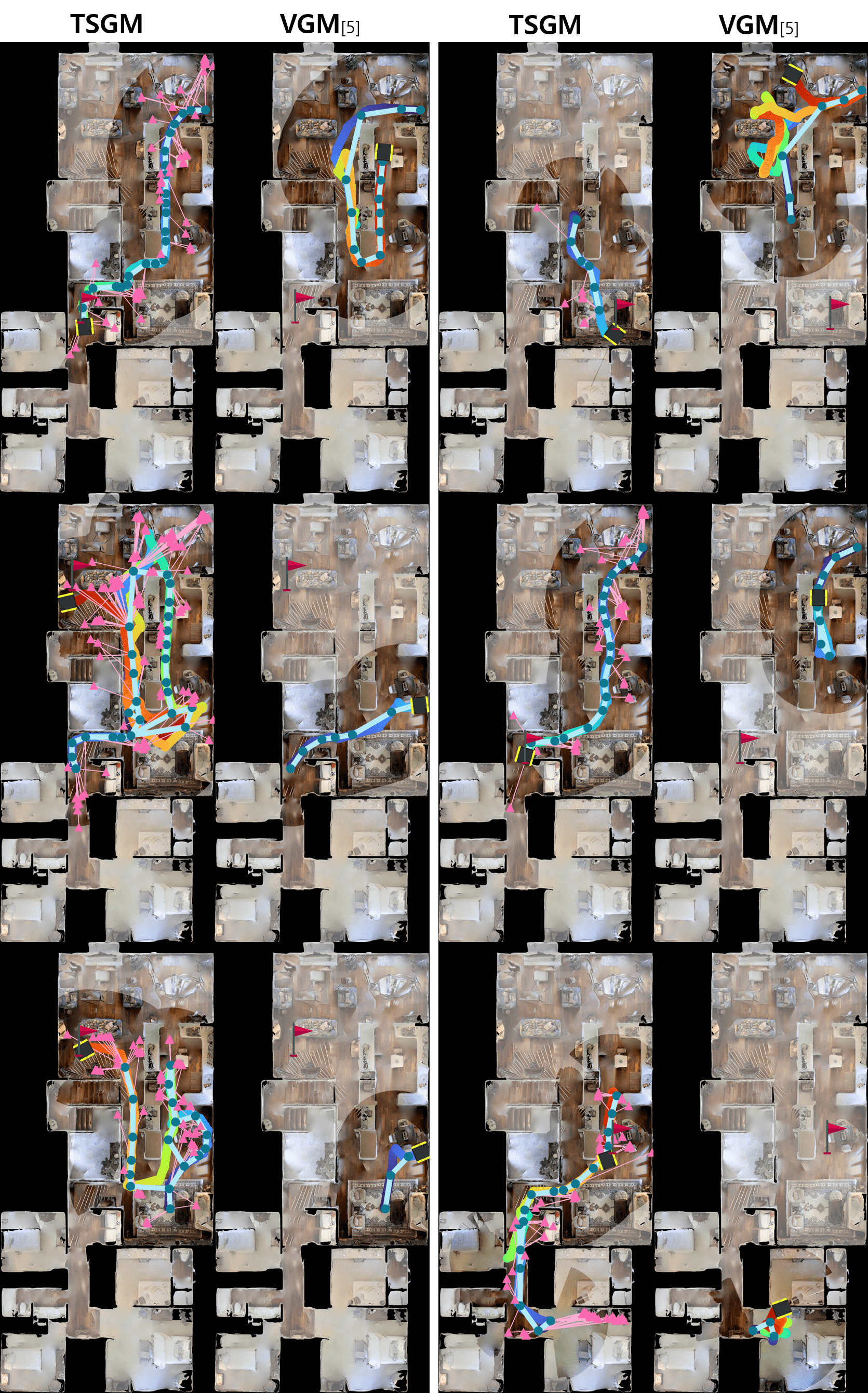}}\centering
\caption{Examples from Gibson's Mosquito environment.}
\label{fig:example6_1}
\end{figure}

\begin{figure}[h]{\centering\includegraphics[width=\linewidth]{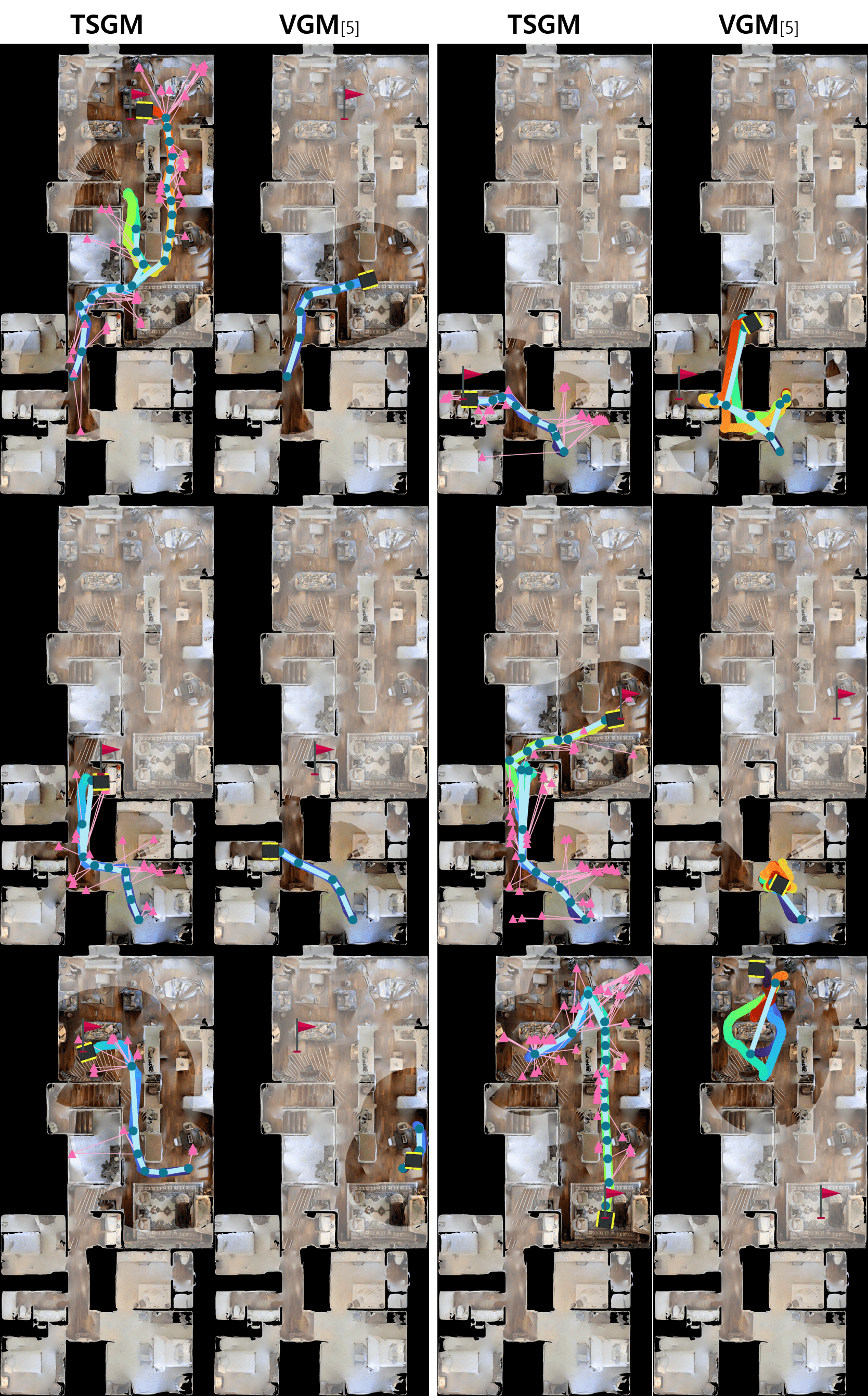}}\centering
\caption{Examples from Gibson's Mosquito environment.}
\label{fig:example6_2}
\end{figure}

\begin{figure}[h]{\centering\includegraphics[width=\linewidth]{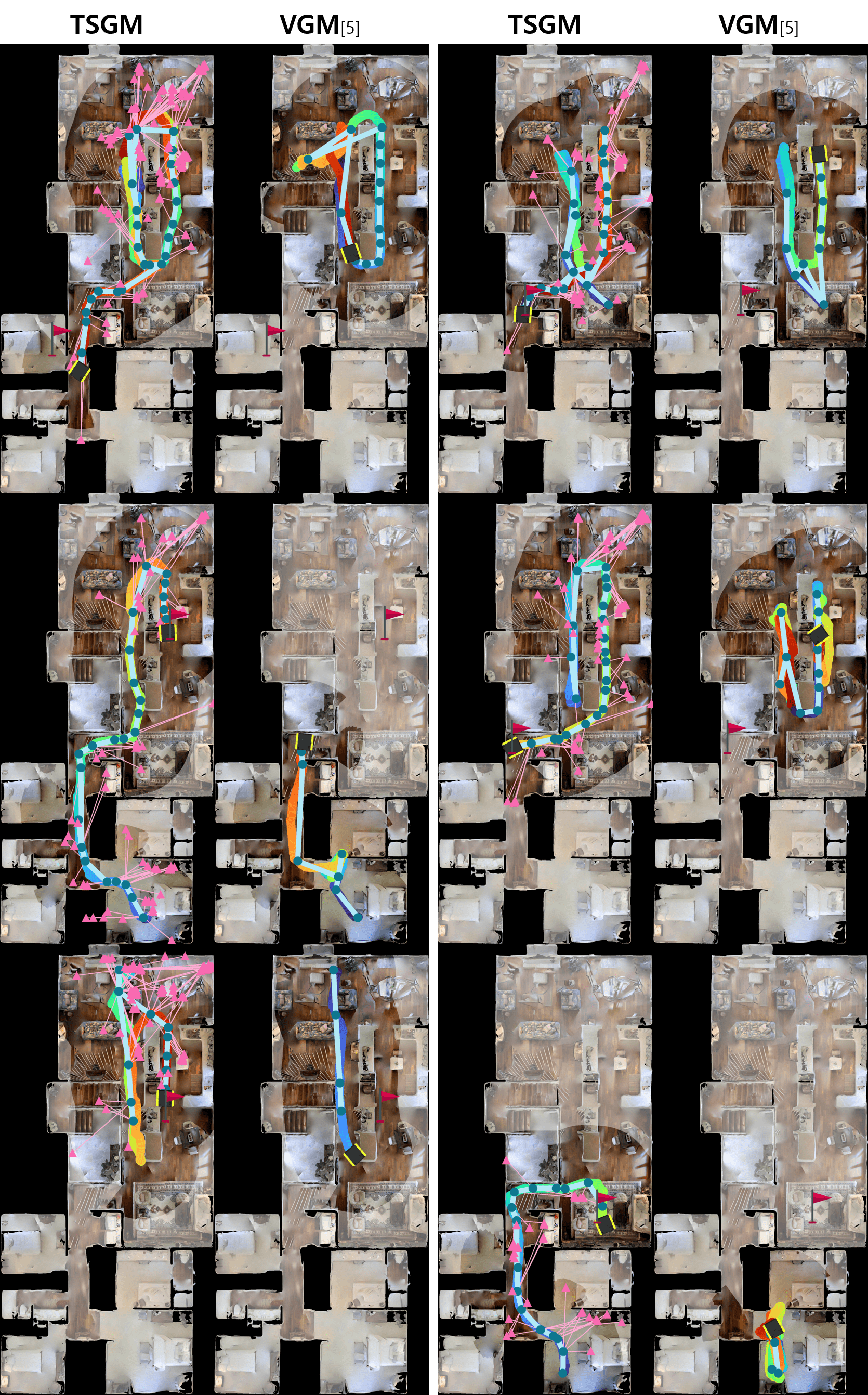}}\centering
\caption{Examples from Gibson's Mosquito environment.}
\label{fig:example6_3}
\end{figure}

\begin{figure}[h]{\centering\includegraphics[width=\linewidth]{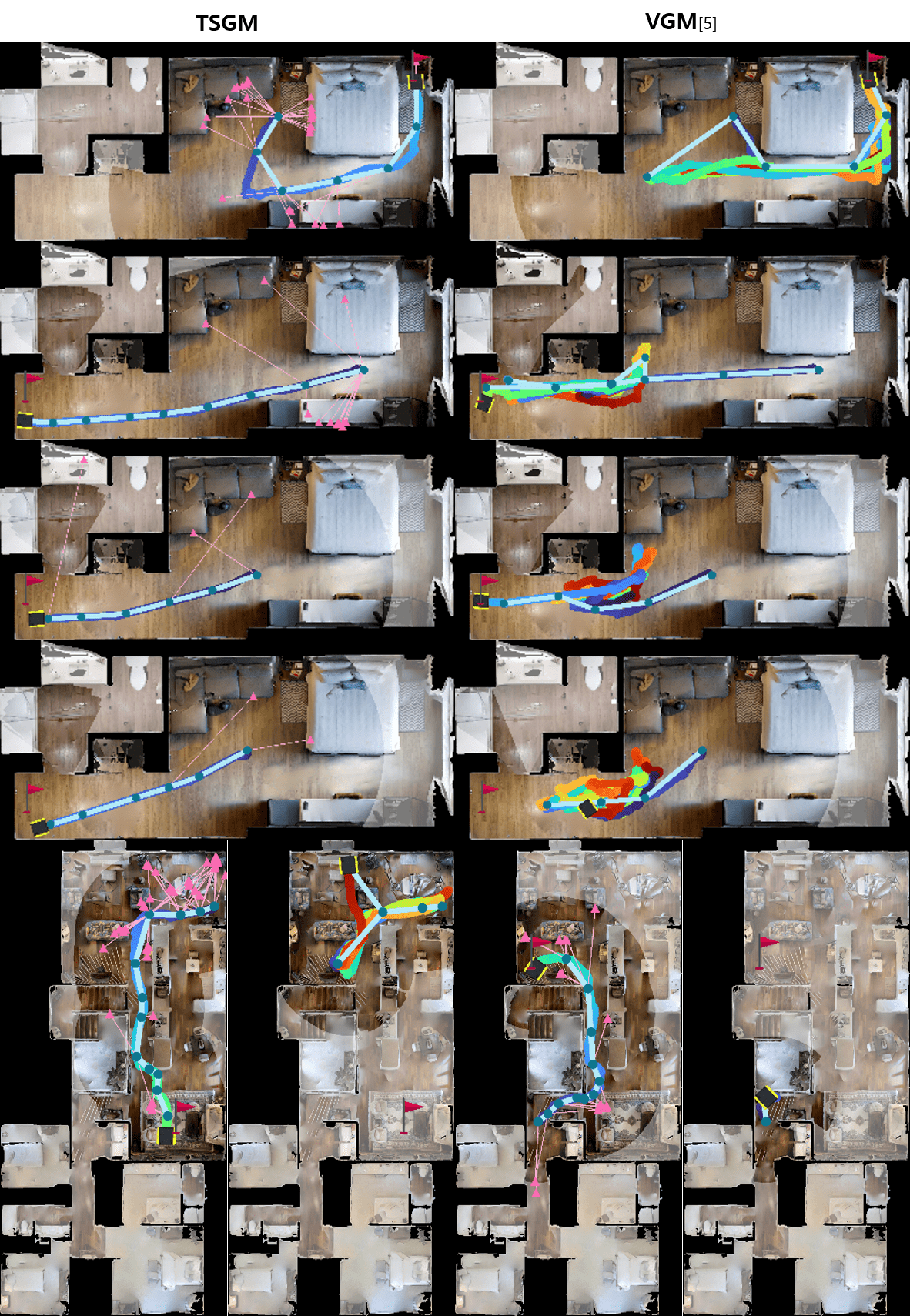}}\centering
\caption{Examples from Gibson's Pablo (top) and Mosquito (bottom) environment.}
\label{fig:example7} 
\end{figure}

\begin{figure}[h]{\centering\includegraphics[width=\linewidth]{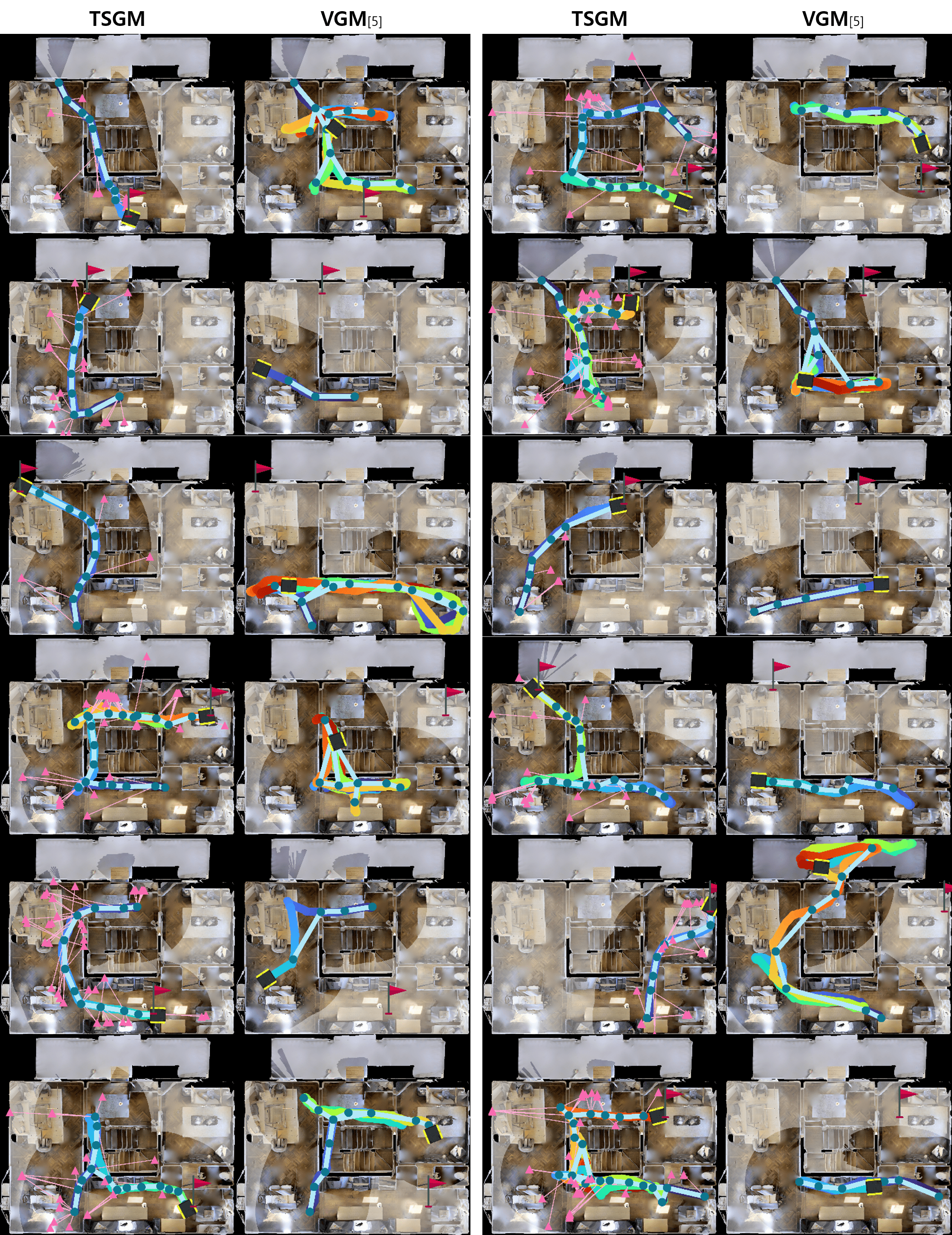}}\centering
\caption{Examples from Gibson's Scioto environment.}
\label{fig:example9_1}
\end{figure}

\begin{figure}[h]{\centering\includegraphics[width=\linewidth]{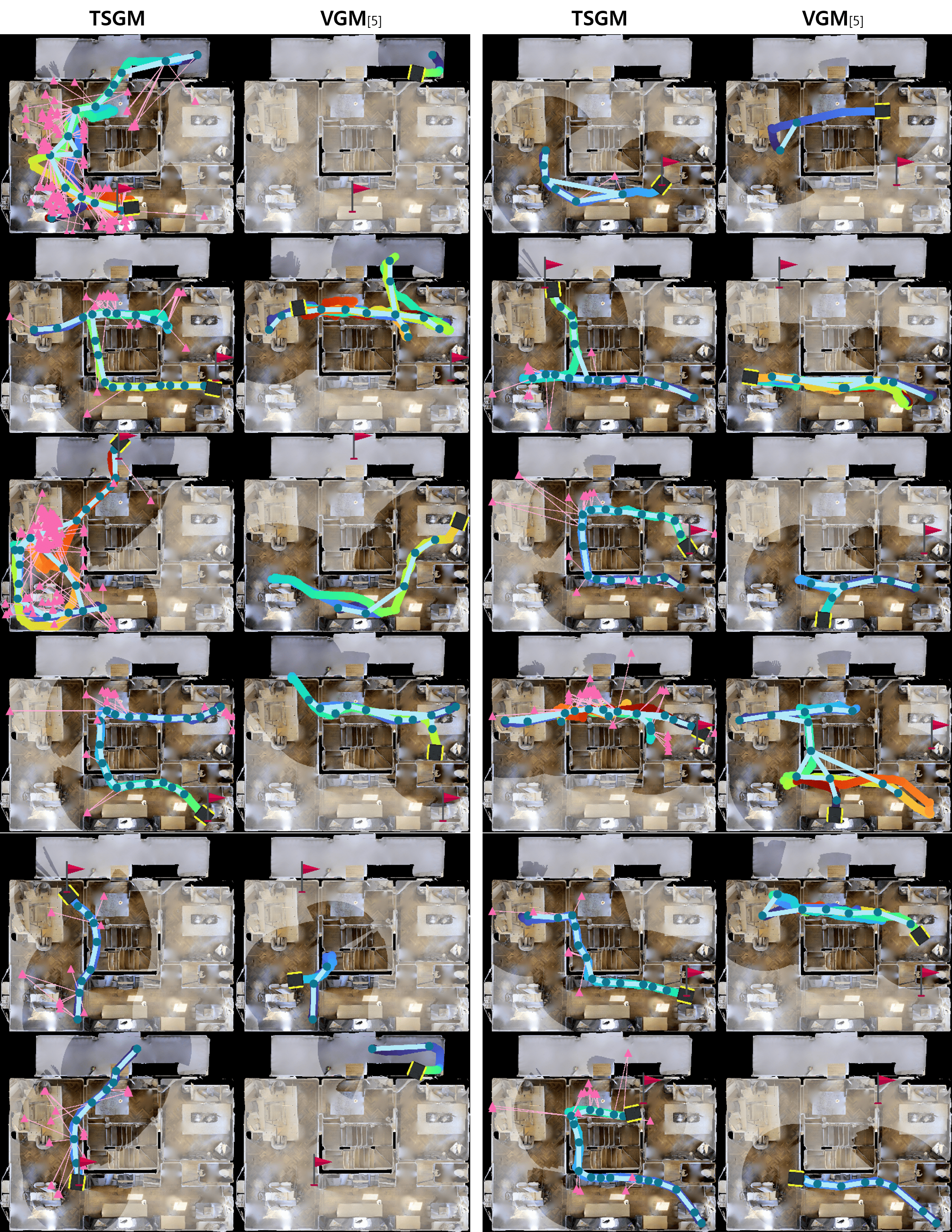}}\centering
\caption{Examples from Gibson's Scioto environment.}
\label{fig:example9_2}
\end{figure}

\begin{figure}[h]{\centering\includegraphics[width=\linewidth]{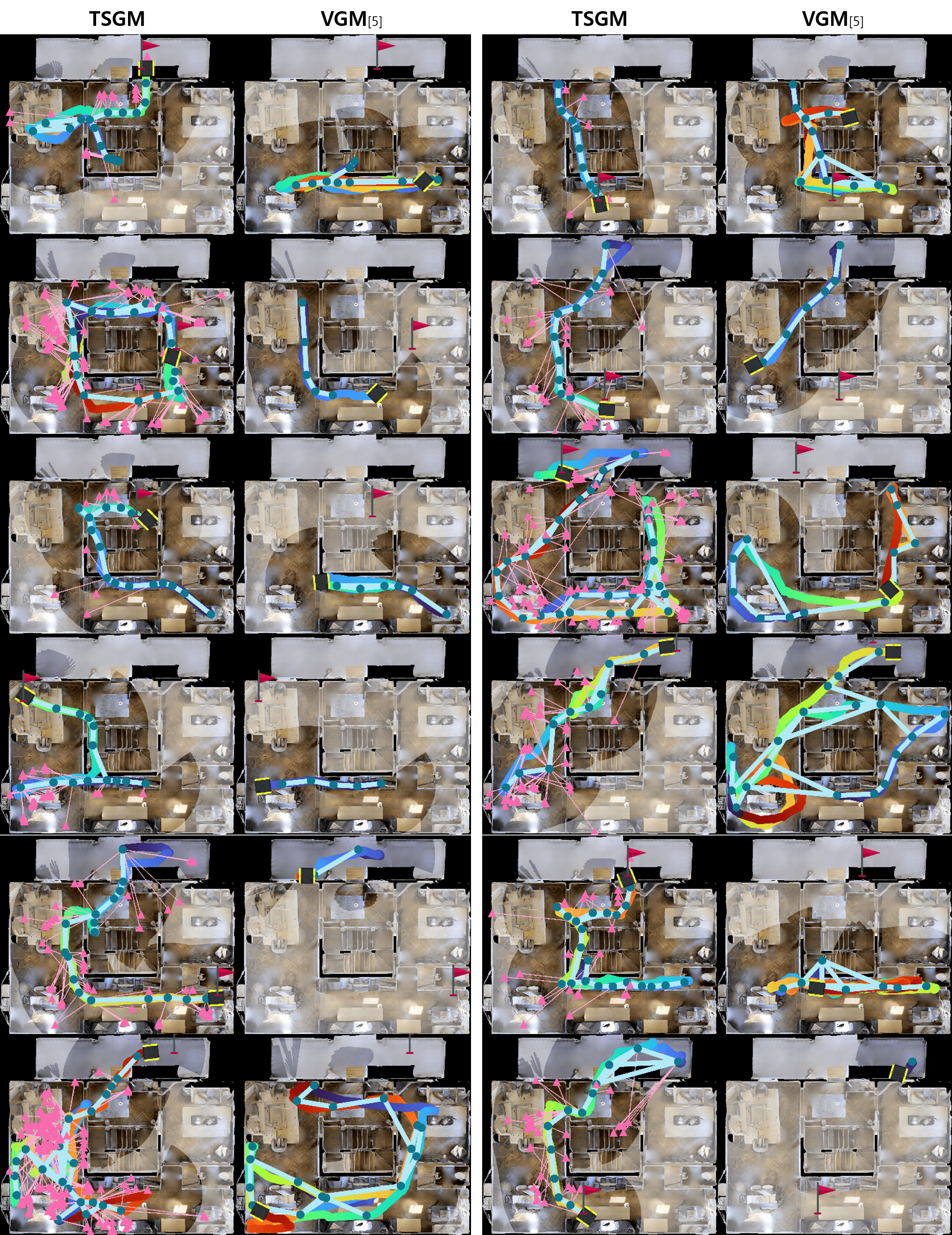}}\centering
\caption{Examples from Gibson's Scioto environment.}
\label{fig:example9_3}
\end{figure}

Robot paths on the simulator are illustrated in Figure~\ref{fig:example1},\ref{fig:example2},\ref{fig:example4},\ref{fig:example5},\ref{fig:example6_1},\ref{fig:example6_2},\ref{fig:example6_3},\ref{fig:example7},\ref{fig:example9_1},\ref{fig:example9_2},\ref{fig:example9_3}.
In the Figures, the color gradation represents the flow of time. The blue arrow symbolizes the starting point, and the red flag indicates the destination. An image node is represented by a blue circle, and an object node is represented by a pink triangle.
The paths are from test episodes proposed in VGM~\cite{VGM}, also used in Table 1.
The paths are drawn in two columns in pairs.
The path generated by the proposed algorithm is in the left column, and the path obtained by the comparison algorithm is in the right column.
The routes provided are more effective and efficient than the comparison algorithm~\cite{VGM} and reach towards the target.

% \begin{figure}[h]{\centering\includegraphics[width=\linewidth]{image/path_example3.png}}\centering
% \caption{Examples from Gibson's Denmark environment.}
% \label{fig:example3}
% \end{figure}
% \begin{figure}[h]{\centering\includegraphics[width=\linewidth]{image/path_example8.png}}\centering
% \caption{Examples from Gibson's Ribera environment.}
% \label{fig:example8}
% \end{figure}
\end{appendices}
\end{document}